\documentclass{article}

\PassOptionsToPackage{numbers,compress}{natbib}

\usepackage[english]{babel}
\usepackage[letterpaper,top=2cm,bottom=2cm,left=3cm,right=3cm,marginparwidth=1.75cm]{geometry}
\usepackage[parfill]{parskip}
\usepackage[utf8]{inputenc}
\usepackage[T1]{fontenc}
\usepackage{microtype}

\usepackage{graphicx}

\usepackage{natbib}

\usepackage{amsmath,amsfonts,bm}

\def\eqref#1{equation~\ref{#1}}
\def\1{\bm{1}}

\DeclareMathAlphabet{\mathsfit}{\encodingdefault}{\sfdefault}{m}{sl}
\SetMathAlphabet{\mathsfit}{bold}{\encodingdefault}{\sfdefault}{bx}{n}

\usepackage{amssymb}
\usepackage{pifont}
\usepackage{booktabs}
\usepackage{subcaption}
\usepackage{xcolor}
\makeatletter
\newcommand{\resultrowshade}{\noalign{\begingroup\color{gray!15}\hrule height\dimexpr\ht\@arstrutbox+\dp\@arstrutbox\relax\kern-\dimexpr\ht\@arstrutbox+\dp\@arstrutbox\relax\endgroup}}
\makeatother
\usepackage{tikz}
\usetikzlibrary{arrows.meta,positioning,calc}
\usepackage{algorithm}
\usepackage{algpseudocode}
\usepackage[colorlinks=true,allcolors=blue]{hyperref}
\usepackage{xurl}
\usepackage{tabularx,array}
\usepackage[most]{tcolorbox}
\usepackage{enumitem}
\usepackage{placeins}
\definecolor{PromptBorder}{HTML}{B9BEC4}
\definecolor{PromptTitle}{HTML}{EEF0F2}
\newtcolorbox{memoryprompt}[2]{%
  enhanced,breakable,sharp corners,
  colback=white,colframe=PromptBorder,colbacktitle=white,
  coltitle=black,boxrule=0.4pt,titlerule=0.3pt,
  left=10pt,right=10pt,top=8pt,bottom=8pt,
  toptitle=5pt,bottomtitle=5pt,
  fonttitle=\small\bfseries,
  title={Prompt #1\quad #2},
  fontupper=\fontsize{10}{12}\selectfont,
  before skip=8pt,after skip=12pt,
  before upper={\setlength{\parindent}{0pt}\setlength{\parskip}{5pt}\raggedright},
}

\newtcolorbox{memoryexample}[2]{%
  enhanced,breakable,sharp corners,
  colback=white,colframe=PromptBorder,colbacktitle=white,
  coltitle=black,boxrule=0.4pt,titlerule=0.3pt,
  left=10pt,right=10pt,top=8pt,bottom=8pt,
  toptitle=5pt,bottomtitle=5pt,
  fonttitle=\small\bfseries,
  title={Example #1\quad #2},
  fontupper=\fontsize{10}{12}\selectfont,
  before skip=8pt,after skip=12pt,
  before upper={\setlength{\parindent}{0pt}\setlength{\parskip}{5pt}\raggedright},
}

\newtcolorbox{memoryinput}[1]{%
  enhanced,breakable,sharp corners,
  colback=white,colframe=PromptBorder,colbacktitle=white,
  coltitle=black,boxrule=0.4pt,titlerule=0.3pt,
  left=10pt,right=10pt,top=7pt,bottom=7pt,
  toptitle=4pt,bottomtitle=4pt,
  fonttitle=\small\bfseries,title={Inputs for Prompt #1},
  fontupper=\ttfamily\fontsize{9}{10.5}\selectfont,
  before skip=7pt,after skip=10pt,
  before upper={\setlength{\parindent}{0pt}\setlength{\parskip}{0pt}\raggedright},
}

\title{Can Test-Time Memory Evolution Outperform \\ Test-Time Training for Scientific Discovery?}
\author{%
  \setlength{\tabcolsep}{7pt}%
  \begin{tabular}{cccc}
    Yepeng Liu$^{*,\dagger}$ & Shengmai Chen$^{*}$ & Xuandong Zhao & Chengzhi Liu \\
    UC Santa Barbara & UC Santa Barbara & UC Berkeley & UC Santa Barbara
  \end{tabular} \\[2ex]
  \setlength{\tabcolsep}{16pt}%
  \begin{tabular}{ccc}
    Dawn Song & Xin Eric Wang & Yuheng Bu \\
    UC Berkeley & UC Santa Barbara & UC Santa Barbara
  \end{tabular}
}

\date{}

\begin{document}
\maketitle

% Manual title-page symbols match the ICWBench arXiv author layout.
{\renewcommand{\thefootnote}{\fnsymbol{footnote}}%
\footnotetext[1]{These authors contributed equally to this work.}%
\footnotetext[2]{Correspondence to: \href{mailto:yepengliu@ucsb.edu}{yepengliu@ucsb.edu}.}}

\begin{abstract}
Recent \emph{test-time training} (TTT) frameworks have enabled open-weight large language models (LLMs) to outperform inference-only frameworks powered by frontier models on scientific discovery tasks. However, TTT introduces additional computational cost for online weight updates during the discovery process, making computational efficiency an important consideration when scaling to larger models. This motivates us to explore an efficient \emph{test-time adaptation} method for scientific discovery. We introduce \emph{Test-Time Memory Evolution} (TTME), a plug-and-play test-time adaptation method, as an alternative to TTT. TTME continually updates the external memory by distilling the experience to guide subsequent discovery without updating model weights. To have a clean comparison, we evaluate TTME and TTT under the same discovery framework, replacing only the test-time adaptation method.  Our evaluation shows that TTME outperforms TTT on all five mathematical tasks, while reducing computational cost by $82.8$-$84.6\%$ in FLOPs and $47.7$-$58.4\%$ in runtime. These results establish TTME as an effective and computationally efficient alternative to TTT in the evaluated discovery settings. Our code is available at \url{https://github.com/yepengliu/TTME}.

% Under the same discovery framework, 
% By controlling the discovery framework, we have a clean comparison between TTME and TTT on both performance and cost.

% We introduce Test-Time Memory Evolution (TTME), a plug-and-play approach that continually updates external memory to guide a fixed-weight model through its input context. TTME distills execution and evaluation feedback into practical lessons for candidate implementation and extracts reusable algorithmic strategies from score-improving attempts. It selects and adapts these strategies for subsequent exploration, enabling ideas from different search attempts to be combined. We replace test-time weight updates with TTME in TTT-Discover and ThetaEvolve and evaluate three models spanning 8B and 120B parameters across five mathematical construction tasks. TTME outperforms the reported TTT results in all nine evaluated configurations. Our cost analysis under TTT-Discover with Qwen3-8B shows 82.8--84.6\% fewer FLOPs and 47.7--58.4\% less runtime excluding evaluation, including all memory-related model calls. These results establish evolving external memory as an effective and computationally efficient alternative to test-time weight adaptation in the evaluated discovery settings.
\end{abstract}

% \vspace{-1em}

\begin{figure}[h]
  \centering
  \includegraphics[width=\linewidth]{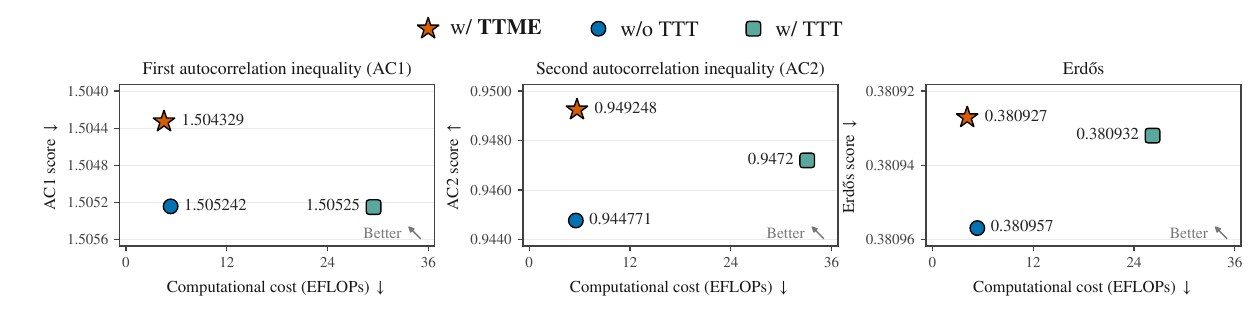}
  % \vspace{-2em}
  % \caption{Best-found scores versus computational cost under the TTT-Discover framework \citep{yuksekgonul2026learning} across three tasks with Qwen3-8B. The three settings use the same discovery framework, with TTT, with TTME replacing TTT, or with neither TTT nor TTME. \textbf{The results demonstrate that \emph{TTME} achieves the best scores at low computational cost on all three tasks.}}
  \caption{Best-found scores versus computational cost under the TTT-Discover framework \citep{yuksekgonul2026learning} with Qwen3-8B. The three settings use the same discovery framework, with TTT, with TTME replacing TTT, or with neither method. \textbf{\emph{TTME} achieves the best scores at low computational cost across all three tasks.}}
  \label{fig:teaser-score-cost}
  % \vspace{-0.5em}
\end{figure}

\section{Introduction}

Scientific discovery often involves proposing hypotheses, evaluation, and using the resulting evidence to guide further investigation \citep{king2009automation,lu2024aiscientist}. Advances in reasoning, coding, and agentic tool-use capabilities of large language models (LLMs) have enabled systems that automate part of this discovery process \citep{novikov2025alphaevolve,lange2025shinkaevolve}. In the iterative discovery loop, a failed attempt can reveal an implementation constraint, while a successful modification can suggest an idea worth applying elsewhere. A key design question is how to learn from this accumulated experience to improve subsequent search efficiently.

% How a discovery system should accumulate and reuse these experiences is therefore central to making progress within a limited search budget.

Recent \emph{test-time training} (TTT) frameworks, such as TTT-Discover \citep{yuksekgonul2026learning} and ThetaEvolve \citep{wang2025thetaevolve}, provide a promising solution to this challenge by combining iterative search with test-time reinforcement learning (RL) \citep{shao2024deepseekmath,deepseekai2025deepseekr1}. Specifically, an LLM generates candidate solutions and receives evaluation feedback. These evaluated attempts provide training data for RL, which internalizes search experience in model weights to guide subsequent generation.
% Specifically, given a discovery problem, an LLM generates candidate solutions and gets feedback from a verifier. These evaluated attempts provide problem-specific training data to update the model's weights through RL. Through TTT, the model internalizes experience accumulated during the search into its weights, which improves the quality of future exploration. 
Their results demonstrate that TTT is a powerful \emph{test-time adaptation} method on challenging scientific discovery tasks, enabling smaller open-weight models to outperform inference-only systems based on frontier models, such as AlphaEvolve \citep{novikov2025alphaevolve}. However, TTT requires online model weight adaptation during the whole discovery process, which is usually computationally expensive and time-consuming (see Table~\ref{tab:efficiency}). It makes TTT hard to scale to larger models for more complex discovery tasks. This motivates us to explore: \emph{under the same discovery framework, is there a more efficient test-time adaptation method that can achieve comparable or even better performance than test-time training?}

% \begin{center}
%     \it under the same discovery framework, is there a more efficient test-time adaptation method that can achieve comparable or even better performance than test-time training?
% \end{center}
% This leaves an important question: \emph{Is there a more efficient alternative to test-time training for scientific discovery?}

% To investigate this problem, we compare test-time adaptation methods within the same discovery framework. Specifically, we study whether updating external memory can provide a more efficient alternative to updating model weights. Our core intuition is that LLMs could learn from their discovery experience through in-context learning, without updating their weights \citep{brown2020language,shinn2023reflexion,zhao2024expel,zhang2026ace}. 

We hypothesize that evolving external memory could be a efficient alternative to online weight updates for scientific discovery. Our core intuition is that current LLMs already have broad foundational knowledge, reasoning and coding capabilities to use experience accumulated during discovery. Continually distilling and organizing this experience in external memory may therefore be sufficient to improve subsequent discovery, without internalizing it through weight updates. Building on this intuition, we introduce Test-Time Memory Evolution (TTME), a plug-and-play \emph{self-evolving memory} system for scientific discovery. We update the memory from evaluation feedback and provide relevant content in the context to guide the subsequent steps.

% Instead of internalizing search experience through updating weights, we extract it from evaluated attempts and store it in external memory for LLMs. We continually update the memory using evaluation feedback and use it to guide the next search step.

% We provide relevant memory content in the model's context to guide subsequent generation.

Specifically, we divide the search experience into two components: practical lessons for implementing candidate solutions and algorithmic ideas for improving task scores. Correspondingly, TTME maintains two complementary memory modules: \emph{Guideline Memory} and \emph{Strategy Memory}. Guideline Memory distills execution and evaluation feedback into shared operational guidelines for reliable implementation. Strategy Memory extracts effective algorithmic ideas from score-improving modifications and consolidates related experiences into reusable strategies. We select and adapt relevant strategies to guide subsequent exploration, combining ideas from different search attempts.

Our contributions are summarized as follows:

\begin{itemize}[leftmargin=1.2em,labelsep=0.4em]
  % \item We investigate whether a self-evolving memory can outperform test-time weight updates in scientific discovery under the same discovery frameworks. 
  % Our study shows that evolving external memory can provide a more effective and computationally efficient alternative to updating model weights.
  % \item We introduce TTME, an efficient plug-and-play method that continually updates external memory from evaluated search experience to guide subsequent discovery with a fixed-weight model.
  \item We investigate whether a test-time memory evolution can outperform test-time weight updates in scientific discovery, and introduce an plug-and-play self-evolving memory method that continually updates external memory from evaluated search experience to guide subsequent discovery with a fixed-weight model.

  \item We evaluate TTME under two TTT frameworks across five mathematical construction tasks and three models spanning 8B and 120B parameters. \emph{We preserve each framework's experimental configuration, replacing only its TTT part with TTME. TTME outperforms the reported TTT results in all evaluated tasks.} Our analysis shows that TTME reduces running time by $47.7-58.4\%$ and computational costs by  $82.8-84.6\%$ relative to TTT. These results demonstrate that TTME is a more efficient alternative to TTT for scientific discovery.
  
  \item We analyze the search behavior of TTME and TTT during the discovery. Both methods improve the implementation validity rate relative to the underlying discovery framework without TTT or TTME, but TTME achieves better performance. Moreover, compared with TTT, TTME more frequently builds on score-improving combinations of previously explored ideas.
  
  %Component ablations support combining both memories, while a two-task factorial study shows that adding TTT to TTME reduces performance.
\end{itemize}

% \begin{itemize}
%   \item \textbf{Rethinking adaptation for scientific discovery.} We investigate a central question: \emph{Can Test-Time Memory Evolution outperform Test-Time Training for scientific discovery?} Our study shows that evolving external memory can provide a more effective and computationally efficient alternative to updating model weights.
%   \item \textbf{Test-Time Memory Evolution.} We introduce TTME, a plug-and-play method that turns discovery experience into evolving guidance for a fixed-weight model. Guideline Memory captures shared operational lessons, while Strategy Memory consolidates algorithmic ideas for adaptation and combination in subsequent search.
%   \item \textbf{Empirical evaluation and analysis.} TTME outperforms the reported TTT results in all nine tested configurations across two frameworks and three models. Our Qwen3-8B cost analysis also shows lower computational cost. Component ablations support combining both memories, while a two-task factorial study shows that adding TTT to TTME reduces performance.
% \end{itemize}

\section{Related Work}

\paragraph{Non-Parametric Adaptation.} Non-parametric discovery methods use fixed-weight LLMs to generate executable candidates and guide subsequent search with evaluation feedback \citep{romeraparedes2024funsearch,si2026towards,lange2025shinkaevolve,liu2026convexbench,liu2026figures,liu2024llm4ad}. The effectiveness of this process depends on how search policies interact with the operators used to generate and modify candidates \citep{toledo2026ai,chen2026avo}. FunSearch \citep{romeraparedes2024funsearch} and AlphaEvolve \citep{novikov2025alphaevolve,georgiev2025mathematical} evolve candidate programs through repeated generation and evaluation with proprietary LLMs. Adaptive search methods further use observed progress to adjust exploration intensity and allocate computation across candidate populations \citep{cemri2026adaevolve}. These studies establish search organization as an important factor in the quality and efficiency of automated discovery. Beyond selecting promising candidates, execution experience can also help subsequent exploration \cite{agrawal2026gepa,zhang2026ace,lyu2026evoscientist,liu2026memevo,shinn2023reflexion,zhao2024expel,yang2024opro}. \cite{ye2024reevo} and \cite{chen2026mars} compare candidate solutions and their evaluation results to identify promising modifications. \cite{cao2026k} uses execution feedback to guide which optimizations to attempt next by adding, reprioritizing, or pruning proposed modifications in a search tree. These studies improve discovery through the joint design of search, planning, and feedback utilization. We instead study experience accumulation specifically as a modular adaptation mechanism and introduce TTME, a plug-and-play memory system for existing discovery frameworks.

\paragraph{Parametric Adaptation.} Parametric methods update the LLMs through RL on evaluated search attempts \citep{surina2025evotune,phan2025migrate,hubert2026alphaproof}. ThetaEvolve \citep{wang2025thetaevolve} turns evolutionary program search into a dynamic RL environment. As its program archive evolves, training proceeds from newly discovered candidates rather than repeatedly starting from the same initial program. TTT-Discover \citep{yuksekgonul2026learning} designs both its learning objective and search rule around finding a single high-quality solution rather than maximizing average performance. It exponentially upweights high-reward attempts during training and uses the best observed outcomes to prioritize states for further search. These methods internalize search experience in model weights, enabling open-weight models to surpass results achieved with proprietary models \citep{wang2025thetaevolve,yuksekgonul2026learning}. However, continually updating model weights during discovery incurs additional computational cost. We show that evolving external memory offers a more effective and efficient alternative to online weight adaptation. TTME achieves better discovery outcomes within both frameworks, and our efficiency experiments demonstrate lower computational cost.

% Preamble dependencies: amsmath, amssymb, tikz, algorithm, algpseudocode.
% TikZ libraries: arrows.meta, positioning, calc.
% Standalone review build: local/methodology_preview.tex.
\section{Method}
\label{sec:method}

\subsection{Background and Problem Formulation}
\label{sec:problem-setup}

We study scientific discovery as an iterative process of proposing hypotheses, evaluating them, and using the resulting evidence to guide subsequent proposals. We focus on tasks with executable candidate programs and a task-specific verifier, where a program produces a construction whose validity and objective value are assessed by the verifier.

Let $v(x)\in\{0,1\}$ indicate whether candidate program $x$ produces a valid result, and let $s(x)$ denote its evaluated score, oriented so that larger values are better. For the set $\mathcal{X}_{\mathrm{eval}}$ of candidates evaluated within the search budget, the best valid score is
\begin{equation}
  s^\star =
  \max_{\substack{x\in\mathcal{X}_{\mathrm{eval}}\\v(x)=1}} s(x).
  \label{eq:discovery-objective}
\end{equation}
Our goal is to find the highest-scoring valid candidate within a given search budget.

Our study builds on two existing TTT scientific discovery frameworks, TTT-Discover \citep{yuksekgonul2026learning} and ThetaEvolve \citep{wang2025thetaevolve}. We describe two frameworks using a common formulation. Each framework maintains an archive of search states, each representing a candidate solution and its associated records. At each search step, the framework selects states from the archive as parents for new attempts. The code associated with a selected parent is its parent program. Given the task specification, parent program, and associated records, the LLM generates candidate programs. We execute and evaluate these programs, recording their validity, scores, and execution feedback. The framework then updates the archive for subsequent search. We preserve each framework’s original parent-selection and archive-update rules.

% A \emph{search state} represents a candidate solution from which search can continue. We use \emph{archive} to denote the collection of retained search states and associated records in either framework. A \emph{parent} is a state selected from this archive as the starting point for a new attempt. At each search step, the framework selects parents and constructs generation contexts from the task, selected states, and associated information. The LLM, then, generates programs. Execution and verification produce candidate outcomes, including validity, objective values, and execution feedback. The framework records these attempts in the evaluated search history and updates the archive with states retained for subsequent search. We preserve each framework's parent-selection and archive-update rules. Below, \emph{parent program} refers to the code associated with a selected parent and supplied for modification.
% In TTT-Discover's mathematical tasks, states are constructions, with generating code available as associated context. ThetaEvolve instead stores programs and their evaluation results.

% We treat two existing TTT frameworks, e.g., TTT-Discover \citep{yuksekgonul2026learning} and ThetaEvolve \citep{wang2025thetaevolve}, as a common search interface.

\begin{figure}[t]
  \centering
  \includegraphics[width=0.9\linewidth]{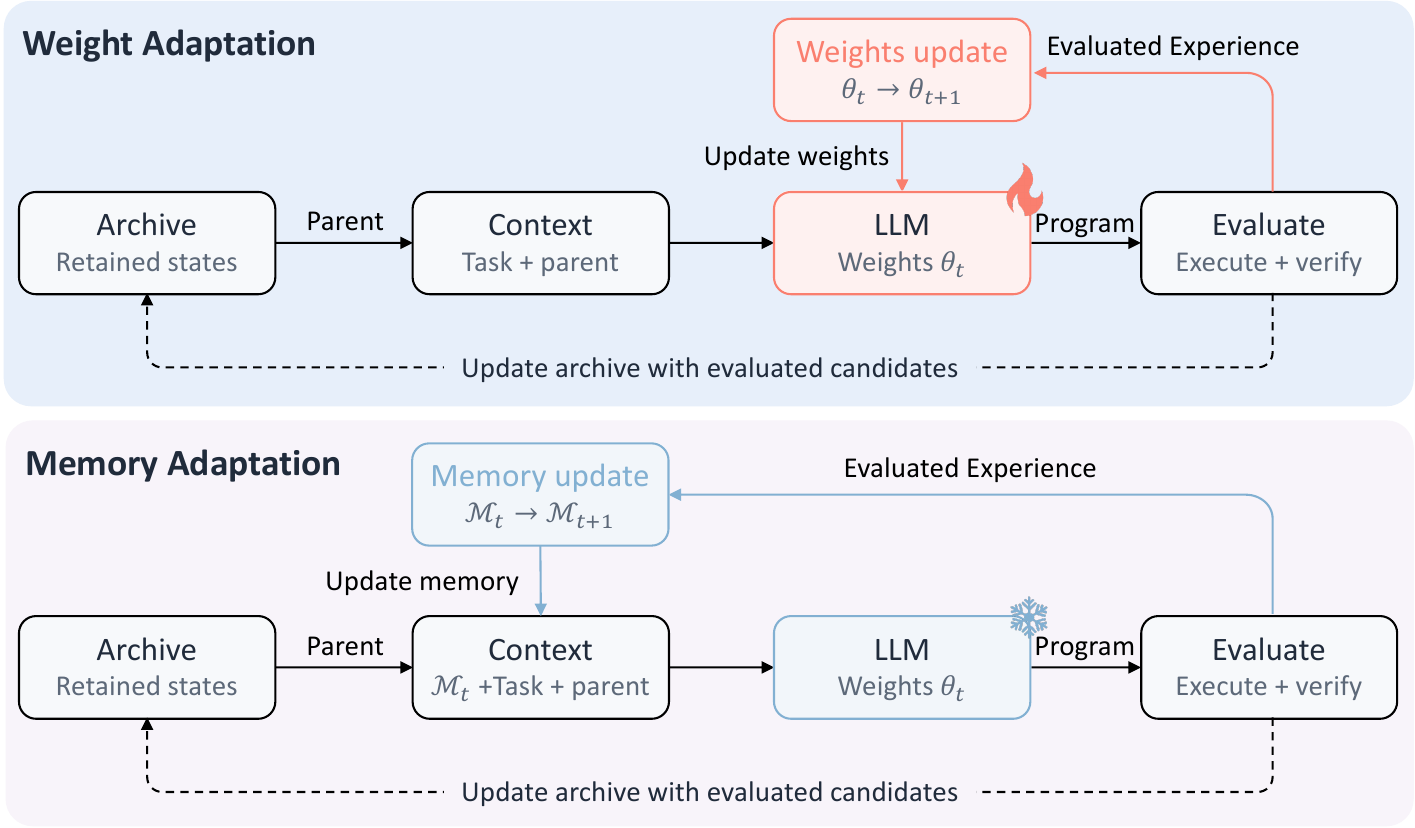}
  \caption{Weight adaptation updates model parameters through test-time RL (top). Memory adaptation updates external memory that conditions generation while keeping model weights fixed (bottom). Both retain the same underlying discovery framework.}
  \label{fig:overview}
  % \vspace{-1em}
\end{figure}

% These search operations already adapt subsequent prompts to the evaluated history. 
Both frameworks use search feedback to update model weights through test-time RL. \emph{We ask whether non-parametric methods, e.g., memory adaptation, can substitute for these parametric methods, e.g., online weight adaptation}. Writing $\mathcal{D}_t$ for evaluated search records available to an update, we contrast two adaptation mechanisms:
\begin{equation}
  \begin{aligned}
    \text{Weight adaptation:}\quad
      &\theta_{t+1} \leftarrow U_{\mathrm{RL}}(\theta_t,\mathcal{D}_t),\\
    \text{Memory adaptation:}\quad
      &\mathcal{M}_{t+1}
        \leftarrow U_{\mathrm{mem}}(\mathcal{M}_t,\mathcal{D}_t),
        \qquad \theta_t=\theta_0.
  \end{aligned}
  \label{eq:adaptation-mechanisms}
\end{equation}
Here, $U_{\mathrm{RL}}$ updates model parameters, whereas $U_{\mathrm{mem}}$ updates external memory used to construct subsequent generation contexts. $t$ denotes the search step, $\theta_t$ is the model parameter, and $\mathcal{M}_t$ is the external memory. We retain the underlying discovery framework and task verifier while replacing only TTT with TTME. Figure~\ref{fig:overview} compares these two adaptation mechanisms.

Evaluated search attempts provide practical lessons for implementing candidate solutions and algorithmic ideas for improving task scores. We organize this experience into two complementary memories, $\mathcal{M}_t=(\mathcal{L}_t,\mathcal{S}_t)$. Guideline Memory $\mathcal{L}_t$ stores shared guidelines for implementing candidate solutions. Strategy Memory $\mathcal{S}_t$ organizes algorithmic ideas into reusable strategies that we select and adapt to guide subsequent search.

% Search records contain both operational feedback and examples of algorithmic improvement. Reusing this experience across parents requires extracting useful changes from their original code and adapting them to the current program. We therefore maintain $\mathcal{M}_t=(\mathcal{L}_t,\mathcal{S}_t)$: \emph{Guideline Memory} $\mathcal{L}_t$ stores shared operational lessons, and \emph{Strategy Memory} $\mathcal{S}_t$ organizes reusable algorithmic experience. We use Strategy Memory to guide search through parent-specific modification instructions, while both generation branches receive the shared guidelines. Together, these memories provide reusable guidance for subsequent search while keeping the model weights fixed. 

% Figure~\ref{fig:overview} illustrates how new outcomes update the guidance for subsequent search.

% These directions are instructions for the current generation step; the two memories persist across steps. The same fixed-weight base model $\pi_\theta$ performs generation and memory-related calls. Figure~\ref{fig:memory-overview} illustrates how new outcomes update the guidance for subsequent search.

\subsection{Guideline Memory: Shared Operational Lessons}
\label{sec:guideline-memory}

Execution and evaluation feedback can reveal implementation problems and choices that improve task scores. Lessons from one attempt can guide the implementation of later candidates. We distill these observations into compact guidelines stored in $\mathcal{L}_t$. Each guideline specifies when it applies, explains the relevant failure or improvement, and recommends a concrete action.

Every $K$ search steps, we form a summary $r_t$ of outcomes collected across parents since the previous update, including valid results, their scores, and observed failures. We update the memory in two stages, first proposing guidelines $\mathcal{G}_t$ and then incorporating them into the existing memory:
\begin{equation}
  \mathcal{G}_t \leftarrow \operatorname{Distill}(r_t,\mathcal{L}_t),
  \qquad
  \mathcal{L}_{t+1} \leftarrow \operatorname{Maintain}(\mathcal{L}_t,\mathcal{G}_t,t).
  \label{eq:guideline-update}
\end{equation}
Here, $\operatorname{Distill}$ prompts the same base model with $r_t$ and the existing guidelines $\mathcal{L}_t$ to propose $\mathcal{G}_t$. In $\operatorname{Maintain}$, we match each proposal against existing entries. For a match, we increment the entry's recurrence count and refresh its last reaffirmed step to $t$; otherwise, we add a new entry. We remove stale entries and enforce a capacity limit, prioritizing recurrence counts and breaking ties by recency. The retained guidelines provide shared context for implementing subsequent candidate solutions.

\subsection{Strategy Memory: Reusable Algorithmic Experience}
\label{sec:strategy-memory}

Combining existing ideas is important in scientific discovery.  An idea behind a score-improving modification may also be useful for other candidates. Strategy Memory preserves these ideas and how they were applied, so they can be adapted and combined in subsequent searches.

% When a hypothesis and its corresponding implementation achieve a higher evaluation score in one parent program, the underlying idea may also be useful for parents in other parts of the search space. Strategy Memory consolidates these experiences into reusable strategies, and records the conditions under which they were applied. 

We update Strategy Memory every $K$ search steps, together with Guideline Memory. We select evaluated cases where a modified program produces a valid result with a higher score than its parent, and denote this collection by $\mathcal{D}_t^{+}$. For each case, we analyze the original and modified programs, their differences, and the evaluation results using the same base model. The model extracts one or more \emph{experience notes} from each case, each describing an algorithmic idea, how it was applied, and its source case. We denote the resulting collection of notes by $\mathcal{E}_t$.

% describing the algorithmic idea reflected in the changes and the conditions under which it was applied.

Different modifications can implement the same algorithmic idea in different ways. We group experience notes that share an underlying idea into a \emph{strategy family} and maintain a persistent \emph{strategy entry} for each family. Each entry describes the shared idea, its application conditions, and the different ways it has been implemented, with references to the source cases. The extraction of experience notes and the update of Strategy Memory are expressed as:
\begin{equation}
  \mathcal{E}_t \leftarrow \operatorname{Extract}(\mathcal{D}_t^{+}),
  \qquad
  \mathcal{S}_{t+1} \leftarrow \operatorname{Consolidate}(\mathcal{S}_t,\mathcal{E}_t).
  \label{eq:strategy-update}
\end{equation}
Here, $\operatorname{Extract}$ denotes the case-wise extraction described above. In $\operatorname{Consolidate}$, we use the same base model to compare new notes with existing entries and assign them to existing or new families. For an existing family, the model combines the new notes with the current entry to update its content. For a new family, the model creates an entry from its notes.

% Different modifications can use the same underlying idea with different implementation details. We group experience notes that share a similar idea into a \emph{strategy family} and maintain a persistent \emph{strategy entry} for each family. The strategy entry consolidates the experience notes into a reusable description of the shared idea, its application conditions, and its implementation variants, with references to the source cases. 

% At each update, we incorporate new experience notes into existing strategy entries or create new entries for distinct ideas, preserving their implementation details and source references. Thus, Strategy Memory accumulates strategies and their observed applications for subsequent search.

% For example, notes that apply the same adaptive step-size rule at different points in an optimization loop belong to one family, whose entry preserves these placement differences.

\subsection{Strategy-Guided Search and Candidate Generation}
\label{sec:memory-generation}

With two memories established, we now describe how they guide new candidate generation. Guideline Memory provides shared guidance for implementation. Strategy Memory contains ideas drawn from specific modifications, which may need adaptation before they can be applied to another parent. We therefore propose \emph{Strategy-Guided Search} to select relevant strategies and turn them into parent-specific modification instructions. When no suitable strategy is selected, we use the parent context and Strategy Memory to propose a new idea and formulate its modification instruction.

% Guideline Memory directly provides shared context for all candidate generation. Strategy Memory stores a collection of ideas, but not all ideas are relevant to every parent. Moreover, reusing an idea requires adapting it to the current parent program. We therefore introduce \emph{Strategy-Guided Search} to select relevant ideas from Strategy Memory and adapt them to the current parent program. When no suitable strategy is selected, we use the memory and parent context to propose a new idea. In either case, we produce a modification instruction describing how to modify the current parent program.

Specifically, at search step $t$, let $p_i^{(t)}$ denote the program associated with the $i$-th selected parent. Let $C_i^{(t)}$ denote the input context supplied by the underlying framework for the $i$-th parent at step $t$, such as task specification and parent program $p_i^{(t)}$. 

% A \emph{parent group} consists of the generation attempts associated with that parent within the step.
% This history provides local evolutionary context, while the memories supply experience accumulated across parents.

\begin{algorithm}[t]
  \caption{Scientific Discovery with Test-Time Memory Evolution}
  \label{alg:memory-search}
  \begin{algorithmic}[1]
    \Require Base model, task verifier, search budget,
             $B$ candidates per parent, $N$ modification instructions per parent,
             memory update interval $K\geq 1$, Direct Generation fraction $\alpha\in[0,1]$.
    \State Initialize archive, evaluated history, and memories
           $\mathcal{L}_0,\mathcal{S}_0$
    \For{each search step $t=0,1,\ldots$ within the budget}
      \State Sample parents using a selection rule from the underlying framework
      \For{each parent $i$, sequentially}
        \State Select up to $N$ distinct strategies using Eq.~(\ref{eq:strategy-selection})
      \EndFor
      \For{each parent $i$}
        \State Form $N$ modification instructions using
               Eq.~(\ref{eq:modification-instruction})
        \State Generate $\alpha B$ direct candidates using
               Eq.~(\ref{eq:direct-generation})
        \State Generate $(1-\alpha)B/N$ candidates per instruction using
               Eq.~(\ref{eq:guided-generation})
      \EndFor
      \State Evaluate candidates; record outcomes and commit archive updates
      \If{$(t+1)\bmod K=0$}
        % \State Update $\mathcal{L}_{t+1}$ from $\mathcal{L}_t$ and evaluated outcomes
        % \State Update $\mathcal{S}_{t+1}$ using $\mathcal{S}_t$ and evaluated improvements
        \State Update $\mathcal{L}_{t+1}$ using Eq.~(\ref{eq:guideline-update})
        \State Update $\mathcal{S}_{t+1}$ using Eq.~(\ref{eq:strategy-update})
      \Else
        \State $\mathcal{L}_{t+1}\gets\mathcal{L}_t$, $\mathcal{S}_{t+1}\gets\mathcal{S}_t$
      \EndIf
    \EndFor
    \State \Return Best valid result found under the task objective
  \end{algorithmic}
\end{algorithm}

For each parent, we use a \emph{selector} (base model) to analyze and select up to $N$ distinct strategies from Strategy Memory based on their potential to improve the parent program. The selector is provided with the parent context, a summary of each strategy entry, and the strategies already assigned to earlier parents in the current step. Selection proceeds sequentially across parents:
\begin{equation}
  \mathbf{a}_i^{(t)}
  = \operatorname{Select}
    \bigl(C_i^{(t)},\hat{\mathcal{S}}_t,
          \mathbf{A}_{<i}^{(t)}\bigr),
  \qquad
  \mathbf{a}_i^{(t)}=(a_{i,1}^{(t)},\ldots,a_{i,N}^{(t)}),
  \label{eq:strategy-selection}
\end{equation}
Here, $\hat{\mathcal{S}}_t$ denotes the summaries of strategy entries. We collect the nonempty strategy selections for earlier parents into
\[
  \mathbf{A}_{<i}^{(t)}
  =
  \left\{
    a_{j,n}^{(t)}
    \;\middle|\;
    1\leq j<i,\;
    1\leq n\leq N,\;
    a_{j,n}^{(t)}\neq\varnothing
  \right\}.
\]
Each nonempty $a_{i,n}^{(t)}$ denotes a selected strategy. If fewer than $N$ strategies are selected, we set the remaining elements of $\mathbf{a}_i^{(t)}$ to $\varnothing$. The selector prioritizes improvement potential for the current parent. Among similarly plausible choices, it prefers a strategy not yet included in $\mathbf{A}_{<i}^{(t)}$.

% , and $a_{i,n}^{(t)}$ is the $n$-th selected strategy or null if no suitable strategy is selected. The selector is instructed to prioritize improvement potential for the current parent. Among similarly plausible choices, it should prefer a strategy not yet assigned in the step. If fewer than $N$ strategies are selected, we set the remaining elements of $\mathbf{a}_i^{(t)}$ to $\varnothing$.

Then, we produce a parent-specific modification instruction $d_{i,n}^{(t)}$ for $n=1,\ldots,N$:
\begin{equation}
  d_{i,n}^{(t)} =
  \begin{cases}
    \operatorname{Adapt}\bigl(C_i^{(t)},\mathcal{S}_t[a_{i,n}^{(t)}]\bigr),
      & a_{i,n}^{(t)}\neq\varnothing,\\
    \operatorname{Propose}\bigl(C_i^{(t)},\mathcal{S}_t\bigr),
      & a_{i,n}^{(t)}=\varnothing.
  \end{cases}
  \label{eq:modification-instruction}
\end{equation}
For a selected strategy, one \emph{adaptor} (base model) reads its full entry and the full parent code, then specifies how to apply the strategy to that parent. When $a_{i,n}^{(t)}=\varnothing$, the adaptor uses the parent and available strategy memory content to propose a new idea. 

%The resulting directions are transient instructions; proposed ideas enter persistent Strategy Memory through subsequent evaluated improvements and mining.

We use the constructed instructions to guide candidate generation. Adapting an explored strategy allows us to combine an idea extracted from a score-improving modification with the parent program's existing approach. Moreover, to explore additional ideas that may improve the task score, we also generate candidates without a modification instruction.

Specifically, let $B$ denote the total candidate budget per parent. Both generation branches receive the parent context $C_i^{(t)}$ and Guideline Memory $\mathcal{L}_t$. We generate $\alpha B$ candidates through \emph{Direct Generation} using this shared context, where $\alpha$ denotes the fraction of the budget allocated to this branch. We set $\alpha=1/2$ in our experiments. For \emph{Strategy-Guided Generation}, we additionally supply a modification instruction $d_{i,n}^{(t)}$ and generate $(1-\alpha)B/N$ candidates for each of the $N$ instructions:
\begin{align}
  x &\leftarrow \operatorname{Generate}(C_i^{(t)},\mathcal{L}_t),
  \label{eq:direct-generation}\\
  x &\leftarrow \operatorname{Generate}(C_i^{(t)},\mathcal{L}_t,d_{i,n}^{(t)}).
  \label{eq:guided-generation}
\end{align}
The candidates' evaluation outcomes inform subsequent memory updates, which provide guidance for later search steps. The complete procedure is summarized in Algorithm~\ref{alg:memory-search}. Implementation details and prompts appear in Appendices~\ref{app:memory-implementation} and~\ref{app:prompts}.

\section{Experiments}

\subsection{Experiment Settings}

\paragraph{Models.} We use Qwen3-8B \citep{qwen3technicalreport} and gpt-oss-120b \citep{openai2025gptoss120b} in the TTT-Discover framework. For ThetaEvolve, we use DeepSeek-R1-0528-Qwen3-8B \citep{deepseekai2025deepseekr1}, abbreviated as Distill-Qwen3-8B. In our method, we keep the model weights fixed throughout the search. The same base model performs candidate generation and all memory-related operations.

\paragraph{Frameworks.} We evaluate our method within two frameworks, TTT-Discover \citep{yuksekgonul2026learning} and ThetaEvolve \citep{wang2025thetaevolve}, that combine iterative search with test-time RL. We compare three settings within two frameworks: w/o TTT, which keeps model weights fixed without our TTME; w/ TTT, which updates model weights through test-time RL; and w/ TTME, which keeps model weights fixed and updates self-evolving memory. For weight adaptation, we report results from the original papers, with model weights updated through test-time RL. Without weight adaptation, we run the underlying search procedure with fixed model weights and its existing generation context. Our method also keeps the weights fixed, but updates Guideline Memory and Strategy Memory to guide subsequent generation. We retain each framework's parent-selection rules, archive-update rules, and task verifier (settings in Appendix~\ref{app:experimental-setup}).

\paragraph{Tasks.} We evaluate five mathematical construction tasks. (1) \texttt{Erd\H{o}s}: Erd\H{o}s's minimum overlap problem searches for density functions that certify tighter upper bounds on the minimum-overlap constant. (2) \texttt{First autocorrelation inequality (AC1)}: AC1 constructs a nonnegative function $f$ to minimize $\|f*f\|_\infty/\|f\|_1^2$, yielding an upper bound on the first autocorrelation constant. Here, $f*f$ denotes the autoconvolution of $f$. (3) \texttt{Second autocorrelation inequality (AC2)}: AC2 constructs a nonnegative function $f$ to maximize ${\|f*f\|_2^2/(\|f*f\|_1\|f*f\|_\infty)}$, yielding a lower bound on the second autocorrelation constant. (4) \texttt{Third autocorrelation inequality (AC3)}: AC3 allows functions with both positive and negative values and seeks a tighter upper bound using the absolute autoconvolution peak. (5) \texttt{Maximal determinant problem (Hadamard)}: Hadamard maximizes the normalized absolute determinant of a $29\times29$ matrix with entries in $\{-1,+1\}$.
% (2) \texttt{First autocorrelation inequality (AC1)}: AC1 constructs nonnegative functions to tighten an upper bound on an inequality constant defined through autoconvolution. (3) \texttt{Second autocorrelation inequality (AC2)}: AC2 constructs nonnegative functions to improve a lower bound on a different autoconvolution-based inequality constant. 

% Under TTT-Discover, we evaluate Erd\H{o}s, AC1, and AC2. Under ThetaEvolve, we evaluate AC2, AC3, and Hadamard. We retain each framework's initialization, parent selection, archive updates, task verifier, and primary candidate budget. Appendix~\ref{app:experimental-setup} details generation and edit-handling settings. We report the best valid task score found during search. Lower is better for Erd\H{o}s, AC1, and AC3; higher is better for AC2 and Hadamard.

% \paragraph{Implementation Details.}

\subsection{Main Results}

\paragraph{Performance across frameworks and models.} TTME outperforms both TTT and search without TTT across two discovery frameworks, three models, and five tasks (Tables~\ref{tab:main-ttt} and~\ref{tab:main-theta}). The comparison with TTT shows that evolving external memory can yield better discovery results than existing weight-update methods. The comparison with search without TTT shows the benefit of adding memory to the underlying fixed-weight search procedure. Together, these results establish TTME as an effective alternative to test-time weight adaptation across the evaluated settings. Appendix~\ref{app:discovered-constructions} visualizes the discovered constructions.

\begin{table}[t]
  \centering
  \caption{Results under the TTT-Discover framework for three settings: w/ TTT, w/o TTT, and w/ TTME. Results for w/ TTT are from \citet{yuksekgonul2026learning} over 50 search steps.}
  \label{tab:main-ttt}
  % \vspace{-1pt}
  \scriptsize
  \setlength{\tabcolsep}{6pt}
  \renewcommand{\arraystretch}{1.12}
  \begin{tabular*}{\linewidth}{@{\extracolsep{\fill}}llrrr@{}}
    \toprule
    Method & Model & Erd\H{o}s $\downarrow$ & AC1 $\downarrow$ & AC2 $\uparrow$ \\
    \midrule
    TTT-Discover (w/ TTT) & Qwen3-8B & 0.380932 & 1.50525 & 0.9472 \\
    TTT-Discover (w/o TTT) & Qwen3-8B & 0.380957 & 1.505242 & 0.944771 \\
    \resultrowshade
    \textbf{Ours (w/ TTME)} & Qwen3-8B & \textbf{0.380927} & \textbf{1.504329} & \textbf{0.949248} \\
    \midrule
    TTT-Discover (w/ TTT) & gpt-oss-120b & 0.380876 & 1.50287 & 0.9591 \\
    TTT-Discover (w/o TTT) & gpt-oss-120b & 0.380945 & 1.505966 & 0.953682 \\
    \resultrowshade
    \textbf{Ours (w/ TTME)} & gpt-oss-120b & \textbf{0.380862} & \textbf{1.502745} & \textbf{0.960784} \\
    \bottomrule
  \end{tabular*}
\end{table}

\begin{table}[t]
  \centering
  \caption{Results under the ThetaEvolve framework for three settings: w/ TTT, w/o TTT, and w/ TTME. Results for w/ TTT are from \citet{wang2025thetaevolve} over 65 search steps.}
  \label{tab:main-theta}
  % \vspace{-1pt}
  \scriptsize
  \setlength{\tabcolsep}{4pt}
  \renewcommand{\arraystretch}{1.12}
  \begin{tabular*}{\linewidth}{@{\extracolsep{\fill}}llrrr@{}}
    \toprule
    Method & Model & AC2 $\uparrow$ & AC3 $\downarrow$ & Hadamard $\uparrow$ \\
    \midrule
    ThetaEvolve (w/ TTT) & Distill-Qwen3-8B & 0.9469 & 1.4930 & 0.5764 \\
    ThetaEvolve (w/o TTT) & Distill-Qwen3-8B & 0.947375 & 1.519277 & 0.571085 \\
    \resultrowshade
    \textbf{Ours (w/ TTME)} & Distill-Qwen3-8B & \textbf{0.948148} & \textbf{1.464224} & \textbf{0.579938} \\
    \bottomrule
  \end{tabular*}
\end{table}

\begin{figure}[t]
  \centering
  \includegraphics[width=\linewidth]{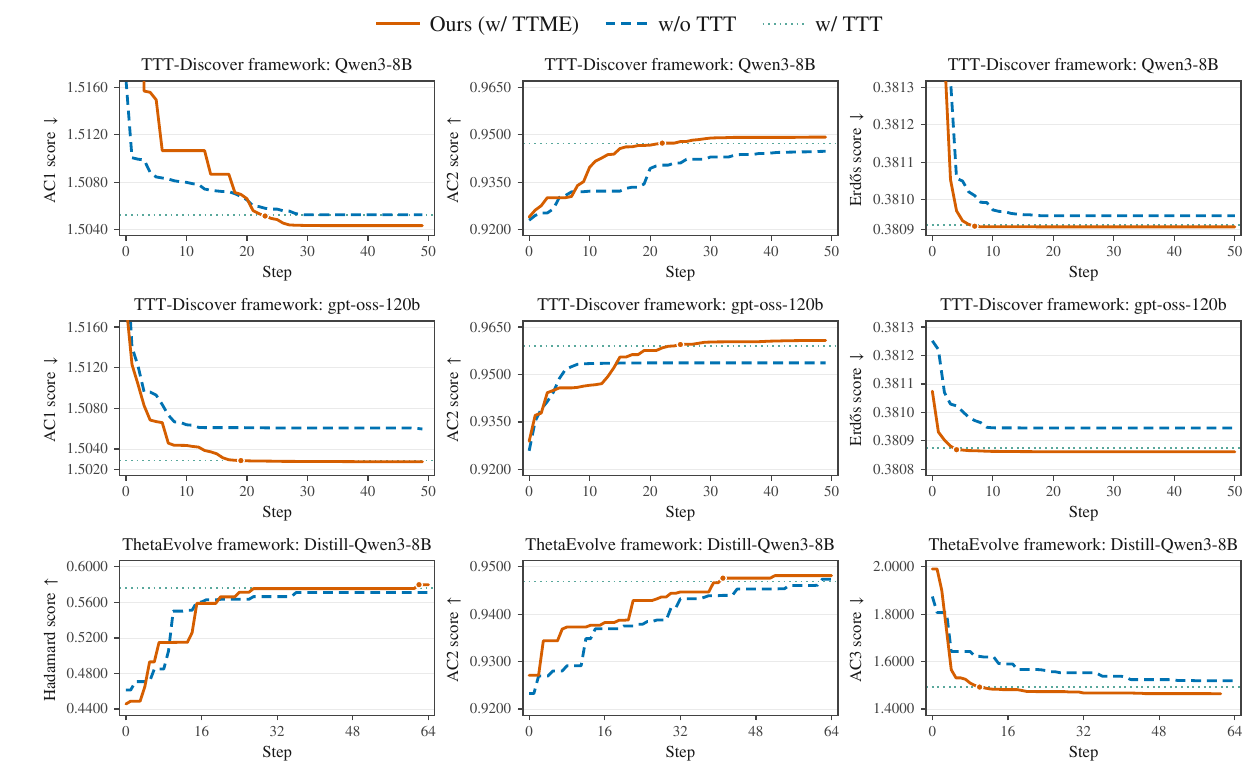}
  % \vspace{-2.5em}
  \caption{Best-found scores during search under TTT-Discover and ThetaEvolve frameworks. TTME achieves better final scores in every evaluated setting. Orange dots mark the first step at which TTME exceeds the corresponding final TTT score.}
  \label{fig:search-progress}
\end{figure}

% Local spacing trial: about 70% of the previous rendered gap.
% \vspace{-9.5pt}
\paragraph{Progress during search.} We next examine how TTME improves the best-found score over the course of search (Figure~\ref{fig:search-progress}). Under the TTT-Discover framework, TTME surpasses the TTT scores within 25 steps across all three tasks and both models. Under the ThetaEvolve framework, TTME also surpasses the TTT scores on three tasks. These trajectories show that TTME can improve on the final TTT results before exhausting the search budget.

% Thus, it exceeds the results of 50-step TTT runs using at most half of the candidate-generation steps.

\begin{table}[H]
  \centering
  \caption{Runtime (excluding evaluation) and computation under TTT-Discover with Qwen3-8B. TTME reduces runtime by 47.7--58.4\% and FLOPs by 82.8--84.6\%.}
  \label{tab:efficiency}
  % \vspace{-1pt}
  \scriptsize
  \setlength{\tabcolsep}{6pt}
  \renewcommand{\arraystretch}{1.12}
  \begin{tabular*}{\linewidth}{@{\extracolsep{\fill}}llrrrrrr@{}}
    \toprule
    & & \multicolumn{3}{c}{Runtime excluding evaluation (h)} & \multicolumn{3}{c}{Computational cost (EFLOPs)} \\
    \cmidrule(lr){3-5}\cmidrule(l){6-8}
    Task & Model & TTT & TTME & Reduction & TTT & TTME & Reduction \\
    \midrule
    AC1 & Qwen3-8B & 11.76 & \textbf{5.36} & 54.5\% & 29.47 & \textbf{4.53} & 84.6\% \\
    AC2 & Qwen3-8B & 12.73 & \textbf{6.66} & 47.7\% & 33.09 & \textbf{5.69} & 82.8\% \\
    Erd\H{o}s & Qwen3-8B & 11.67 & \textbf{4.85} & 58.4\% & 26.21 & \textbf{4.15} & 84.2\% \\
    \bottomrule
  \end{tabular*}
\end{table}

\paragraph{Runtime and computational cost.} We compare runtime (excluding evaluation) and computational cost under TTT-Discover with Qwen3-8B (Table~\ref{tab:efficiency}). We include all candidate generation, memory-related LLM calls, and RL training where applicable. Across the three tasks, even though TTT-Discover updates only LoRA parameters \citep{hu2021lora}, TTME reduces runtime by 47.7--58.4\% and FLOPs by 82.8--84.6\% relative to TTT. Combined with the performance results, these reductions show that TTME achieves better discovery outcomes with less runtime and computation.

% \vspace{-5em}

\subsection{Ablation Studies}

\paragraph{Combining TTME with TTT.} Having established the effectiveness of TTME with fixed model weights, we examine whether adding TTT further improves performance. We compare four settings under the TTT-Discover framework with Qwen3-8B: neither TTT nor TTME, TTT alone, TTME alone, and their combination (Table~\ref{tab:memory-rl-2x2}). In the combined setting, the RL-updated model performs both candidate generation and memory-related operations. TTME alone achieves the best scores on AC2 and Erd\H{o}s, while adding TTT worsens both results. This comparison further supports using TTME as an alternative to TTT rather than requiring both forms of adaptation.

\begin{table}[h]
  \centering
  \caption{Comparison of TTT, TTME, and their combination against a baseline with neither. TTME alone achieves the best scores on both tasks; adding TTT reduces performance.}
  \label{tab:memory-rl-2x2}
  % \vspace{-1pt}
  \scriptsize
  \setlength{\tabcolsep}{6pt}
  \renewcommand{\arraystretch}{1.12}
  \begin{tabular*}{\linewidth}{@{\extracolsep{\fill}}ccrr@{}}
    \toprule
    TTT & TTME & AC2 $\uparrow$ & Erd\H{o}s $\downarrow$ \\
    \midrule
    \multicolumn{4}{@{}l}{\textit{TTT-Discover framework: Qwen3-8B}} \\
    \ding{55} & \ding{55} & 0.944771 & 0.380957 \\
    \ding{51} & \ding{55} & 0.9472 & 0.380932 \\
    \resultrowshade
    \ding{55} & \ding{51} & \textbf{0.949248} & \textbf{0.380927} \\
    \ding{51} & \ding{51} & 0.944141 & 0.380944 \\
    \bottomrule
  \end{tabular*}
\end{table}

\begin{table}[h]
  \centering
  \caption{Ablation of Guideline Memory and Strategy Memory. Full TTME achieves the best scores on all three tasks, demonstrating the benefit of combining both memory components.}
  \label{tab:memory-ablation}
  % \vspace{-1pt}
  \scriptsize
  \setlength{\tabcolsep}{6pt}
  \renewcommand{\arraystretch}{1.12}
  \begin{tabular*}{\linewidth}{@{\extracolsep{\fill}}lrrr@{}}
    \toprule
    Configuration & AC1 $\downarrow$ & AC2 $\uparrow$ & Erd\H{o}s $\downarrow$ \\
    \midrule
    \multicolumn{4}{@{}l}{\textit{TTT-Discover framework: gpt-oss-120b}} \\
    \resultrowshade
    \textbf{Full TTME} & \textbf{1.502745} & \textbf{0.960784} & \textbf{0.380862} \\
    w/o Guideline Memory & 1.505644 & 0.958506 & 0.380865 \\
    w/o Strategy Memory & 1.505981 & 0.956493 & 0.380864 \\
    w/o TTME & 1.505966 & 0.953682 & 0.380946 \\
    \bottomrule
  \end{tabular*}
\end{table}

\paragraph{Contributions of the two memories.} We examine the contributions of Guideline Memory and Strategy Memory under the TTT-Discover framework with gpt-oss-120b (Table~\ref{tab:memory-ablation}). We compare full TTME with variants that remove either memory component or both. Full TTME achieves the best scores on all three tasks, and removing either component worsens performance. These results support retaining both shared operational lessons and reusable algorithmic strategies to guide discovery.

% Keep ablation floats within the ablation subsection.
\FloatBarrier
\subsection{Search Behavior Analysis}

\begin{figure}[h]
  \centering
  \includegraphics[width=\linewidth]{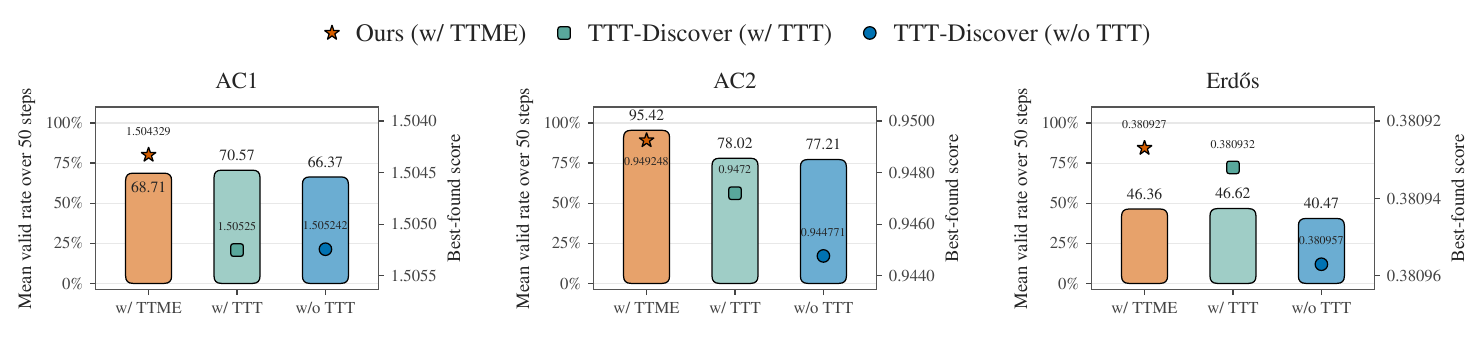}
  % \vspace{-2.5em}
  \caption{Implementation validity rate and final performance under the TTT-Discover framework with Qwen3-8B across three tasks. Bars show mean validity rates over 50 steps (left axes); markers show best-found scores (right axes), with better scores plotted higher.}
  \label{fig:validity-performance}
\end{figure}

\begin{samepage}
\paragraph{Candidate validity and discovery performance.} Successful discovery requires candidates that are both valid and capable of improving the task score. We consider a program valid if it compiles successfully and receives a score from the evaluator. Figure~\ref{fig:validity-performance} compares mean validity over 50 steps and best-found scores under the TTT-Discover framework with Qwen3-8B. Both TTT and TTME improve the mean program validity rate over search without TTT.  On AC1 and Erd\H{o}s, TTT and TTME achieve similar validity rates, while TTME achieves better best-found scores. This suggests TTME generates more effective ideas for improving task scores.
\par
\end{samepage}

% On AC2, TTME achieves both higher validity and a better score than TTT. On AC1, TTME has lower validity but a better score; on Erd\H{o}s, validity is similar while TTME achieves a better score. Candidate validity alone therefore does not capture the quality of discovery outcomes.

\FloatBarrier
\begin{samepage}
\paragraph{Combining ideas for further search.} Beyond validity, we examine how search combines and builds on previously explored ideas under the TTT-Discover framework with Qwen3-8B. We count a candidate as a combination if it retains an idea from its parent and adds an idea previously explored outside its ancestry. The added idea must be absent from the parent. We measure how often these combinations are selected as parents for further search.
\par
\end{samepage}

% , pooling 392 parent selections over steps 1 through 49.

\begin{figure}[h]
  \centering
  \includegraphics[width=0.8\linewidth]{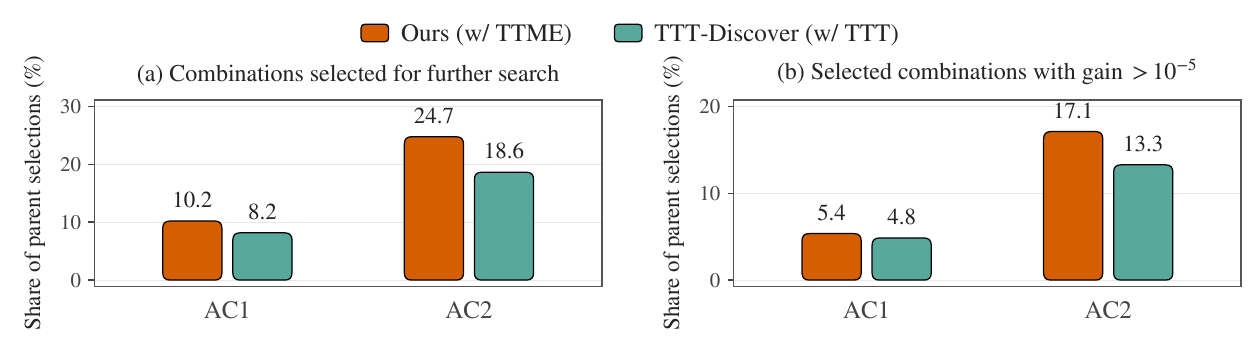}
  % \vspace{-1em}
  \caption{Combination of ideas in subsequent search under TTT-Discover with Qwen3-8B. TTME has a higher combination share on both tasks, with and without the $10^{-5}$ improvement requirement.}
  % TTME not only selects more candidates that combine existing ideas for further exploration, but also more frequently selects such candidates that improve task scores.
  \label{fig:combination-search}
\end{figure}

TTME selects combinations more frequently than TTT on both AC1 (10.2\% versus 8.2\%) and AC2 (24.7\% versus 18.6\%; Figure~\ref{fig:combination-search}). We also count selections of combinations that improve on their own parents by more than $10^{-5}$. TTME again has higher shares on AC1 (5.4\% versus 4.8\%) and AC2 (17.1\% versus 13.3\%). Thus, TTME not only selects combinations of existing ideas more frequently for further search, but also more frequently selects combinations that improve task scores. These observations support using external memory to combine ideas from different search branches and carry productive combinations into subsequent exploration.

\section{Conclusion}

We investigate whether evolving external memory can outperform test-time training for scientific discovery within the same discovery frameworks. We introduce TTME, which distills search experience into operational lessons and algorithmic strategies to guide a fixed-weight model. Across two frameworks, three models, and five mathematical tasks, TTME achieves better scores than TTT in every evaluated setting. Under TTT-Discover with Qwen3-8B, it reduces runtime excluding evaluation by 47.7--58.4\% and FLOPs by 82.8--84.6\%, including all memory-related calls. These results establish self-evolving memory as an effective and computationally efficient alternative to test-time weight updates for the discovery tasks we study.

\FloatBarrier

\clearpage
% \subsection*{Reproducibility statement}
% Section~\ref{sec:method} and Algorithm~\ref{alg:memory-search} describe the
% method and search procedure. Appendix~\ref{app:task-definitions} defines the
% tasks and evaluation metrics. Appendix~\ref{app:experimental-setup} reports
% framework-specific settings, candidate budgets, and upstream code revisions.
% Appendix~\ref{app:memory-implementation} documents memory updates and
% hyperparameters, and Appendix~\ref{app:prompts} provides the memory-operation
% prompts and input templates. Tables~\ref{tab:main-ttt} and~\ref{tab:main-theta}
% identify the published TTT results used for comparison.

\bibliography{iclr2027_conference}

@article{yuksekgonul2026learning,
  title = {Learning to Discover at Test Time},
  author = {Yuksekgonul, Mert and Koceja, Daniel and Li, Xinhao and
            Bianchi, Federico and McCaleb, Jed and Wang, Xiaolong and
            Kautz, Jan and Choi, Yejin and Zou, James and Guestrin, Carlos
            and Sun, Yu},
  journal = {arXiv preprint arXiv:2601.16175},
  year = {2026}
}

@article{wang2025thetaevolve,
  title = {{ThetaEvolve}: Test-time Learning on Open Problems},
  author = {Wang, Yiping and Su, Shao-Rong and Zeng, Zhiyuan and Xu, Eva
            and Ren, Liliang and Yang, Xinyu and Huang, Zeyi and He, Xuehai
            and Ma, Luyao and Peng, Baolin and Cheng, Hao and He, Pengcheng
            and Chen, Weizhu and Wang, Shuohang and Du, Simon Shaolei
            and Shen, Yelong},
  journal = {arXiv preprint arXiv:2511.23473},
  year = {2025}
}

@misc{qwen3technicalreport,
      title={{Qwen3} Technical Report}, 
      author={{Qwen Team}},
      year={2025},
      eprint={2505.09388},
      archivePrefix={arXiv},
      primaryClass={cs.CL},
      url={https://arxiv.org/abs/2505.09388}, 
}

@misc{openai2025gptoss120b,
      title={{gpt-oss-120b} \& {gpt-oss-20b} Model Card}, 
      author={OpenAI},
      year={2025},
      eprint={2508.10925},
      archivePrefix={arXiv},
      primaryClass={cs.CL},
      url={https://arxiv.org/abs/2508.10925}, 
}

@misc{deepseekai2025deepseekr1,
      title={{DeepSeek-R1}: Incentivizing Reasoning Capability in {LLMs} via Reinforcement Learning}, 
      author={DeepSeek-AI},
      year={2025},
      eprint={2501.12948},
      archivePrefix={arXiv},
      primaryClass={cs.CL},
      url={https://arxiv.org/abs/2501.12948}, 
}

@article{romeraparedes2024funsearch,
  title = {Mathematical discoveries from program search with large language models},
  author = {Romera-Paredes, Bernardino and Barekatain, Mohammadamin and Novikov, Alexander and Balog, Matej and Kumar, M. Pawan and Dupont, Emilien and Ruiz, Francisco J. R. and Ellenberg, Jordan S. and Wang, Pengming and Fawzi, Omar and Kohli, Pushmeet and Fawzi, Alhussein},
  journal = {Nature},
  volume = {625},
  number = {7995},
  pages = {468--475},
  year = {2024},
  doi = {10.1038/s41586-023-06924-6},
  url = {https://www.nature.com/articles/s41586-023-06924-6}
}

@article{novikov2025alphaevolve,
  title = {{AlphaEvolve}: A coding agent for scientific and algorithmic discovery},
  author = {Novikov, Alexander and V{\~u}, Ng{\^a}n and Eisenberger, Marvin and Dupont, Emilien and Huang, Po-Sen and Wagner, Adam Zsolt and Shirobokov, Sergey and Kozlovskii, Borislav and Ruiz, Francisco J. R. and Mehrabian, Abbas and Kumar, M. Pawan and See, Abigail and Chaudhuri, Swarat and Holland, George and Davies, Alex and Nowozin, Sebastian and Kohli, Pushmeet and Balog, Matej},
  journal = {arXiv preprint arXiv:2506.13131},
  year = {2025},
  url = {https://arxiv.org/abs/2506.13131}
}

@inproceedings{ye2024reevo,
  title = {{ReEvo}: Large Language Models as Hyper-Heuristics with Reflective Evolution},
  author = {Ye, Haoran and Wang, Jiarui and Cao, Zhiguang and Berto, Federico and Hua, Chuanbo and Kim, Haeyeon and Park, Jinkyoo and Song, Guojie},
  booktitle = {Advances in Neural Information Processing Systems},
  volume = {37},
  pages = {43571--43608},
  year = {2024},
  url = {https://proceedings.neurips.cc/paper_files/paper/2024/file/4ced59d480e07d290b6f29fc8798f195-Paper-Conference.pdf}
}

@article{shinn2023reflexion,
  title = {{Reflexion}: Language Agents with Verbal Reinforcement Learning},
  author = {Shinn, Noah and Cassano, Federico and Berman, Edward and Gopinath, Ashwin and Narasimhan, Karthik and Yao, Shunyu},
  journal = {arXiv preprint arXiv:2303.11366},
  year = {2023},
  url = {https://arxiv.org/abs/2303.11366}
}

@inproceedings{zhao2024expel,
  title = {{ExpeL}: {LLM} Agents Are Experiential Learners},
  author = {Zhao, Andrew and Huang, Daniel and Xu, Quentin and Lin, Matthieu and Liu, Yong-Jin and Huang, Gao},
  booktitle = {Proceedings of the AAAI Conference on Artificial Intelligence},
  year = {2024},
  url = {https://arxiv.org/abs/2308.10144}
}

@inproceedings{zhang2026ace,
  title = {Agentic Context Engineering: Evolving Contexts for Self-Improving Language Models},
  author = {Zhang, Qizheng and Hu, Changran and Upasani, Shubhangi and Ma, Boyuan and Hong, Fenglu and Kamanuru, Vamsidhar and Rainton, Jay and Wu, Chen and Ji, Mengmeng and Li, Hanchen and Thakker, Urmish and Zou, James and Olukotun, Kunle},
  booktitle = {International Conference on Learning Representations},
  year = {2026},
  url = {https://arxiv.org/abs/2510.04618}
}

@inproceedings{agrawal2026gepa,
  title = {{GEPA}: Reflective Prompt Evolution Can Outperform Reinforcement Learning},
  author = {Agrawal, Lakshya A. and Tan, Shangyin and Soylu, Dilara and Ziems, Noah and Khare, Rishi and Opsahl-Ong, Krista and Singhvi, Arnav and Shandilya, Herumb and Ryan, Michael J. and Jiang, Meng and Potts, Christopher and Sen, Koushik and Dimakis, Alexandros G. and Stoica, Ion and Klein, Dan and Zaharia, Matei and Khattab, Omar},
  booktitle = {International Conference on Learning Representations},
  year = {2026},
  url = {https://arxiv.org/abs/2507.19457}
}

@article{lyu2026evoscientist,
  title = {{EvoScientist}: Towards Multi-Agent Evolving {AI} Scientists for End-to-End Scientific Discovery},
  author = {Lyu, Yougang and Zhang, Xi and Yi, Xinhao and Zhao, Yuyue and Guo, Shuyu and Hu, Wenxiang and Piotrowski, Jan and Kaliski, Jakub and Urbani, Jacopo and Meng, Zaiqiao and Zhou, Lun and Yan, Xiaohui},
  journal = {arXiv preprint arXiv:2603.08127},
  year = {2026},
  url = {https://arxiv.org/abs/2603.08127}
}

@article{liu2026memevo,
  title = {{$\varepsilon$-MemEvo}: Adaptive Cross-Task Memory Transfer for {LLM} Program Evolution},
  author = {Liu, Aofan and Song, Shiyuan and Qi, Yiyan},
  journal = {arXiv preprint arXiv:2608.12522},
  year = {2026},
  url = {https://arxiv.org/abs/2608.12522}
}

@article{matolcsi2010autoconvolutions,
  title   = {Improved bounds on the supremum of autoconvolutions},
  author  = {Matolcsi, M{\'a}t{\'e} and Vinuesa, Carlos},
  journal = {Journal of Mathematical Analysis and Applications},
  volume  = {372},
  number  = {2},
  pages   = {439--447.e2},
  year    = {2010},
  doi     = {10.1016/j.jmaa.2010.07.030},
  url     = {https://doi.org/10.1016/j.jmaa.2010.07.030}
}

@article{barnard2020convolution,
  title   = {Three convolution inequalities on the real line with connections
             to additive combinatorics},
  author  = {Barnard, Richard C. and Steinerberger, Stefan},
  journal = {Journal of Number Theory},
  volume  = {207},
  pages   = {42--55},
  year    = {2020},
  doi     = {10.1016/j.jnt.2019.07.001},
  url     = {https://arxiv.org/abs/1903.08731}
}

@article{boyer2026autoconvolution,
  title   = {An improved example for an autoconvolution inequality},
  author  = {Boyer, Christopher and Li, Zane Kun},
  journal = {Experimental Mathematics},
  pages   = {1--7},
  year    = {2026},
  doi     = {10.1080/10586458.2025.2607423},
  url     = {https://doi.org/10.1080/10586458.2025.2607423}
}

@article{haugland2016overlap,
  title   = {The minimum overlap problem revisited},
  author  = {Haugland, Jan Kristian},
  journal = {arXiv preprint arXiv:1609.08000},
  year    = {2016},
  url     = {https://arxiv.org/abs/1609.08000}
}

@article{orrick2003determinant,
  title   = {New lower bounds for the maximal determinant problem},
  author  = {Orrick, William P. and Solomon, Bruce and Dowdeswell, Roland
             and Smith, Warren D.},
  journal = {arXiv preprint arXiv:math/0304410},
  year    = {2003},
  url     = {https://arxiv.org/abs/math/0304410}
}

@article{hu2021lora,
  title={Lora: Low-rank adaptation of large language models},
  author={Hu, Edward J and Shen, Yelong and Wallis, Phillip and Allen-Zhu, Zeyuan and Li, Yuanzhi and Wang, Shean and Wang, Lu and Chen, Weizhu},
  journal={arXiv preprint arXiv:2106.09685},
  year={2021}
}

@article{lange2025shinkaevolve,
  title = {{ShinkaEvolve}: Towards Open-Ended And Sample-Efficient Program Evolution},
  author = {Lange, Robert Tjarko and Imajuku, Yuki and Cetin, Edoardo},
  journal = {arXiv preprint arXiv:2509.19349},
  year = {2025},
  url = {https://arxiv.org/abs/2509.19349}
}

@article{liu2024llm4ad,
  title = {{LLM4AD}: A Platform for Algorithm Design with Large Language Model},
  author = {Liu, Fei and Zhang, Rui and Xie, Zhuoliang and Sun, Rui and Li, Kai and Lin, Xi and Wang, Zhenkun and Lu, Zhichao and Zhang, Qingfu},
  journal = {arXiv preprint arXiv:2412.17287},
  year = {2024},
  url = {https://arxiv.org/abs/2412.17287v1}
}

@article{georgiev2025mathematical,
  title = {Mathematical Exploration and Discovery at Scale},
  author = {Georgiev, Bogdan and G{\'o}mez-Serrano, Javier and Tao, Terence and Wagner, Adam Zsolt},
  journal = {arXiv preprint arXiv:2511.02864},
  year = {2025},
  url = {https://arxiv.org/abs/2511.02864}
}

@article{hubert2026alphaproof,
  title = {Olympiad-level Formal Mathematical Reasoning with Reinforcement Learning},
  author = {Hubert, Thomas and Mehta, Rishi and Sartran, Laurent and
            Horv{\'a}th, Mikl{\'o}s Z. and {\v{Z}}u{\v{z}}i{\'c}, Goran and
            Wieser, Eric and Huang, Aja and Schrittwieser, Julian and
            Schroecker, Yannick and Masoom, Hussain and Bertolli, Ottavia and
            Zahavy, Tom and Mandhane, Amol and Yung, Jessica and
            Beloshapka, Iuliya and Ibarz, Borja and Veeriah, Vivek and
            Yu, Lei and Nash, Oliver and Lezeau, Paul and Mercuri, Salvatore and
            S{\"o}nne, Calle and Mehta, Bhavik and Davies, Alex and Zheng, Daniel and
            Pedregosa, Fabian and Li, Yin and von Glehn, Ingrid and
            Rowland, Mark and Albanie, Samuel and Velingker, Ameya and
            Schmitt, Simon and Lockhart, Edward and Hughes, Edward and
            Michalewski, Henryk and Sonnerat, Nicolas and Hassabis, Demis and
            Kohli, Pushmeet and Silver, David},
  journal = {Nature},
  volume = {651},
  number = {8106},
  pages = {607--613},
  year = {2026},
  doi = {10.1038/s41586-025-09833-y},
  url = {https://www.nature.com/articles/s41586-025-09833-y}
}

@article{phan2025migrate,
  title = {{MiGrATe}: Mixed-Policy {GRPO} for Adaptation at Test-Time},
  author = {Phan, Peter and Agarwal, Dhruv and Srinivas, Kavitha and Samulowitz, Horst and Kapanipathi, Pavan and McCallum, Andrew},
  journal = {arXiv preprint arXiv:2508.08641},
  year = {2025},
  url = {https://arxiv.org/abs/2508.08641}
}

@article{surina2025evotune,
  title = {Algorithm Discovery With {LLMs}: Evolutionary Search Meets Reinforcement Learning},
  author = {Surina, Anja and Mansouri, Amin and Quaedvlieg, Lars and Seddas, Amal and Viazovska, Maryna and Abbe, Emmanuel and Gulcehre, Caglar},
  journal = {arXiv preprint arXiv:2504.05108},
  year = {2025},
  url = {https://arxiv.org/abs/2504.05108}
}

@article{shao2024deepseekmath,
  title = {{DeepSeekMath}: Pushing the Limits of Mathematical Reasoning in Open Language Models},
  author = {Shao, Zhihong and Wang, Peiyi and Zhu, Qihao and Xu, Runxin and Song, Junxiao and Bi, Xiao and Zhang, Haowei and Zhang, Mingchuan and Li, Y. K. and Wu, Y. and Guo, Daya},
  journal = {arXiv preprint arXiv:2402.03300},
  year = {2024},
  url = {https://arxiv.org/abs/2402.03300}
}

@article{king2009automation,
  title = {The Automation of Science},
  author = {King, Ross D. and Rowland, Jem and Oliver, Stephen G. and Young, Michael and Aubrey, Wayne and Byrne, Emma and Liakata, Maria and Markham, Magdalena and Pir, Pinar and Soldatova, Larisa N. and Sparkes, Andrew and Whelan, Kenneth E. and Clare, Amanda},
  journal = {Science},
  volume = {324},
  number = {5923},
  pages = {85--89},
  year = {2009},
  doi = {10.1126/science.1165620},
  url = {https://doi.org/10.1126/science.1165620}
}

@article{lu2024aiscientist,
  title = {The {AI Scientist}: Towards Fully Automated Open-Ended Scientific Discovery},
  author = {Lu, Chris and Lu, Cong and Lange, Robert Tjarko and Foerster, Jakob and Clune, Jeff and Ha, David},
  journal = {arXiv preprint arXiv:2408.06292},
  year = {2024},
  url = {https://arxiv.org/abs/2408.06292}
}

@inproceedings{yang2024opro,
  title = {Large Language Models as Optimizers},
  author = {Yang, Chengrun and Wang, Xuezhi and Lu, Yifeng and Liu, Hanxiao and Le, Quoc V. and Zhou, Denny and Chen, Xinyun},
  booktitle = {International Conference on Learning Representations},
  pages = {12028--12068},
  year = {2024},
  url = {https://proceedings.iclr.cc/paper_files/paper/2024/hash/3339f19c5fcee3ad74502947a32be9e6-Abstract-Conference.html}
}

@article{liu2026convexbench,
  title={ConvexBench: Can LLMs Recognize Convex Functions?},
  author={Liu, Yepeng and Huang, Yu and Wang, Yu-Xiang and Liang, Yingbin and Bu, Yuheng},
  journal={arXiv preprint arXiv:2602.01075},
  year={2026}
}

@article{liu2026figures,
  title={Figures as Programs: Recursive Generation of Editable Scientific Figures},
  author={Liu, Yepeng and Dai, Dasen and Liu, Chengzhi and Song, Yiren and Ci, Hai and Zhang, Yu and Zhang, Qi and Shou, Mike Zheng and Wang, Xin Eric and Bu, Yuheng},
  journal={arXiv preprint arXiv:2609.01006},
  year={2026}
}

@article{si2026towards,
  title={Towards execution-grounded automated AI research},
  author={Si, Chenglei and Yang, Zitong and Choi, Yejin and Cand{\`e}s, Emmanuel and Yang, Diyi and Hashimoto, Tatsunori},
  journal={arXiv preprint arXiv:2601.14525},
  year={2026}
}

@article{chen2026mars,
  title={Mars: Modular agent with reflective search for automated ai research},
  author={Chen, Jiefeng and Mishra, Bhavana Dalvi and Nam, Jaehyun and Meng, Rui and Pfister, Tomas and Yoon, Jinsung},
  journal={arXiv preprint arXiv:2602.02660},
  year={2026}
}

@article{toledo2026ai,
  title={Ai research agents for machine learning: Search, exploration, and generalization in mle-bench},
  author={Toledo, Edan and Hambardzumyan, Karen and Josifoski, Martin and Hazra, Rishi and Baldwin, Nicolas and Audran-Reiss, Alexis and Kuchnik, Michael and Magka, Despoina and Jiang, Minqi and Lupidi, Alisia and others},
  journal={Advances in Neural Information Processing Systems},
  volume={38},
  pages={35309--35348},
  year={2026}
}

@article{cao2026k,
  title={K-search: Llm kernel generation via co-evolving intrinsic world model},
  author={Cao, Shiyi and Mao, Ziming and Gonzalez, Joseph E and Stoica, Ion},
  journal={arXiv preprint arXiv:2602.19128},
  year={2026}
}

@article{cemri2026adaevolve,
  title={Adaevolve: Adaptive llm driven zeroth-order optimization},
  author={Cemri, Mert and Agrawal, Shubham and Gupta, Akshat and Liu, Shu and Cheng, Audrey and Mang, Qiuyang and Naren, Ashwin and Erdogan, Lutfi Eren and Sen, Koushik and Zaharia, Matei and others},
  journal={arXiv preprint arXiv:2602.20133},
  year={2026}
}

@article{chen2026avo,
  title={AVO: Agentic variation operators for autonomous evolutionary search},
  author={Chen, Terry and Ye, Zhifan and Xu, Bing and Ye, Zihao and Liu, Timmy and Hassani, Ali and Chen, Tianqi and Kerr, Andrew and Wu, Haicheng and Xu, Yang and others},
  journal={arXiv preprint arXiv:2603.24517},
  year={2026}
}
\bibliographystyle{unsrtnat}

\clearpage
\appendix
\section*{Appendix}

\section{Task Definitions and Evaluation}
\label{app:task-definitions}

We evaluate five mathematical construction tasks from
TTT-Discover~\citep{yuksekgonul2026learning} and
ThetaEvolve~\citep{wang2025thetaevolve}: three autocorrelation inequalities
(AC1--AC3), Erd\H{o}s' minimum overlap problem, and the
maximal determinant problem (Hadamard).
The scores below retain the task's original objective direction;
Section~\ref{sec:problem-setup} uses a higher-is-better convention for $s(x)$.

\subsection{First Autocorrelation Inequality (AC1)}
For a nonzero, nonnegative step function $f$ supported on
$[-\tfrac14,\tfrac14]$, its autoconvolution is
\[
    (f*f)(t)=\int_{\mathbb R} f(x)\,f(t-x)\,dx.
\]
AC1 minimizes the ratio~\citep{matolcsi2010autoconvolutions,barnard2020convolution}
\begin{equation}
    S_{\mathrm{AC1}}(f)
    = \frac{\lVert f*f\rVert_{\infty}}{\lVert f\rVert_1^2}.
\end{equation}
For heights $a_0,\ldots,a_{n-1}$ on $n$ equal-width intervals, this score is
\begin{equation}
    S_{\mathrm{AC1}}(a)
    = \frac{2n\,\max_k(a*a)_k}{\left(\sum_{i=0}^{n-1}a_i\right)^2},
\end{equation}
where $a*a$ denotes discrete convolution.

\subsection{Second Autocorrelation Inequality (AC2)}
For the same class of step functions, AC2 maximizes the
ratio~\citep{matolcsi2010autoconvolutions,boyer2026autoconvolution}
\begin{equation}
    S_{\mathrm{AC2}}(f)
    = \frac{\lVert f*f\rVert_2^2}
    {\lVert f*f\rVert_1\lVert f*f\rVert_{\infty}}.
\end{equation}
We compute the squared $L^2$ norm by integrating over each linear segment
of the autoconvolution. A segment of width $\Delta$ with endpoint values
$u$ and $v$ contributes
\[
    \frac{\Delta}{3}\bigl(u^2+uv+v^2\bigr).
\]

\subsection{Third Autocorrelation Inequality (AC3)}
AC3 optimizes over real-valued step functions $f$ satisfying
\[
    \operatorname{supp}(f)\subseteq
    \left[-\tfrac14,\tfrac14\right],
    \qquad
    \int_{-1/4}^{1/4} f(x)\,dx\neq0.
\]
The heights may be positive or negative. The objective is to
minimize~\citep{wang2025thetaevolve}
\begin{equation}
    S_{\mathrm{AC3}}(f)
    = \frac{\lVert f*f\rVert_{\infty}}
    {\left(\int_{-1/4}^{1/4}f(x)\,dx\right)^2}.
\end{equation}
For heights $a_0,\ldots,a_{n-1}$ on $n$ equal-width intervals, the score is
\begin{equation}
    S_{\mathrm{AC3}}(a)
    = \frac{2n\,\max_k\lvert(a*a)_k\rvert}
    {\left(\sum_{i=0}^{n-1}a_i\right)^2}.
\end{equation}

\clearpage
\subsection{Erd\H{o}s' Minimum Overlap Problem}
The continuous formulation~\citep{haugland2016overlap} uses a step function
$h$ and its complement $g$:
\[
    \begin{aligned}
        h&:[-1,1]\rightarrow[0,1],
        &\qquad \int_{-1}^{1}h(x)\,dx&=1,\\[3pt]
        g(x)&=1-h(x),
        & x&\in[-1,1].
    \end{aligned}
\]
We extend both functions by zero outside this interval and minimize
the maximum overlap
\begin{equation}
    S_{\text{Erd\H{o}s}}(h)
    = \sup_{t\in\mathbb R}\int_{\mathbb R}h(x)g(x+t)\,dx.
\end{equation}
For heights $h_0,\ldots,h_{n-1}$ on $n$ equal-width intervals, the
constraints are
\[
    0\leq h_i\leq1\quad(i=0,\ldots,n-1),
    \qquad
    \sum_{i=0}^{n-1}h_i=\frac{n}{2}.
\]
The score is
\begin{equation}
    S_{\text{Erd\H{o}s}}(h)
    = \frac{2}{n}\max_{k=-(n-1),\ldots,n-1}
      \sum_{\substack{0\leq i<n\\0\leq i+k<n}}
      h_i(1-h_{i+k}).
\end{equation}

\medskip
\subsection{Maximal Determinant Problem (Hadamard)}
The maximal determinant problem searches for a matrix with the largest
absolute determinant~\citep{orrick2003determinant}. We use the search space
\[
    H\in\{-1,+1\}^{n\times n},\qquad n=29.
\]
We maximize the normalized score used in
ThetaEvolve~\citep{wang2025thetaevolve}:
\begin{equation}
    S_{\mathrm{Had}}(H)
    = \frac{\lvert\det H\rvert}{2^{28}\cdot7^{12}\cdot342}.
\end{equation}
The evaluator computes the determinant using exact integer arithmetic.

\begin{table}[ht]
    \centering
    \scriptsize
    \caption{Evaluation metrics. Arrows indicate the direction of improvement.
    We compute best-found scores over all valid candidates evaluated up to
    the current search step.}
    \label{tab:task-evaluation}
    \renewcommand{\arraystretch}{1.15}
    \begin{tabular*}{0.9\linewidth}{@{\extracolsep{\fill}}lll@{}}
        \toprule
        Task & Score & Best-found \\
        \midrule
        AC1 ($\downarrow$) & Upper bound & Minimum \\
        AC2 ($\uparrow$) & Lower bound & Maximum \\
        AC3 ($\downarrow$) & Upper bound & Minimum \\
        Erd\H{o}s ($\downarrow$) & Upper bound & Minimum \\
        Hadamard ($\uparrow$) & Normalized determinant & Maximum \\
        \bottomrule
    \end{tabular*}
\end{table}

\clearpage

\section{Experimental Setup}
\label{app:experimental-setup}

Within each framework, w/ TTME and w/o TTT use the same fixed-weight model,
task initialization, evaluator, and primary candidate budget. Each run
maintains its own search archive. Section~\ref{app:memory-Overview}
defines the memory conditions.

\subsection{ThetaEvolve Framework}
\label{app:theta-setup}

We use the ThetaEvolve~\citep{wang2025thetaevolve} inference framework
on AC2, AC3, and the $n=29$ Hadamard maximal-determinant problem, with
DeepSeek-R1-0528-Qwen3-8B (Distill-Qwen3-8B) frozen throughout.

\paragraph{Search and generation.}
\label{app:theta-search}

At each step, we select 32 parents from the MAP-Elites archive with
islands and generate 16 primary candidates per parent. The w/ TTME setting allocates
eight candidates to each generation branch. We construct prompts from
the released expert context, parent program, and \texttt{SEARCH/REPLACE}
rules, then update the archive with the evaluation results.
Table~\ref{tab:theta-setup} compares settings with the released
inference-only configuration.

Calls for memory maintenance, strategy selection, and strategy adaptation
add to the 33,280 primary candidate attempts.

\begin{table}[ht]
    \centering
    \small
    \caption{Settings for fixed-weight Distill-Qwen3-8B on AC2, AC3, and Hadamard. Original denotes ThetaEvolve's 65-step early inference-only setting~\citep{wang2025thetaevolve}. Counts are per task and seed.}
    \label{tab:theta-setup}
    \setlength{\tabcolsep}{4pt}
    \renewcommand{\arraystretch}{1.08}
    \begin{tabularx}{\linewidth}{@{}>{\raggedright\arraybackslash}X
        >{\centering\arraybackslash}p{0.20\linewidth}
        >{\centering\arraybackslash}p{0.165\linewidth}
        >{\centering\arraybackslash}p{0.165\linewidth}@{}}
        \toprule
        Setting & Original ThetaEvolve & \multicolumn{2}{c}{Ours} \\
        \cmidrule(l){3-4}
        & w/o TTT & w/o TTT & w/ TTME \\
        \midrule
        \multicolumn{4}{@{}l}{\textit{Search budget and archive}} \\
        Search steps & 65 & 65 & 65 \\
        Parents $\times$ candidates per parent & $32\times16=512$ & $32\times16=512$ & $32\times16=512$ \\
        Primary candidates per run & 33,280 & 33,280 & 33,280 \\
        Random seeds & 42, 1234, 3407 & 3407 & 3407 \\
        Population / islands / archive & 10,000 / 10 / 1,000 & 10,000 / 10 / 1,000 & 10,000 / 10 / 1,000 \\
        \midrule
        \multicolumn{4}{@{}l}{\textit{Generation}} \\
        Separate thinking-token cap & --- & $\times$ & $\times$ \\
        Output format & {\scriptsize\texttt{SEARCH/REPLACE}} & {\scriptsize\texttt{SEARCH/REPLACE}} & {\scriptsize\texttt{SEARCH/REPLACE}} \\
        \midrule
        \multicolumn{4}{@{}l}{\textit{Evaluation}} \\
        Runtime / eval. timeout (s): AC2, Hadamard & 300 / 350 & 300 / 350 & 300 / 350 \\
        Runtime / eval. timeout (s): AC3 & 60 / 70 & 60 / 70 & 60 / 70 \\
        \midrule
        \multicolumn{4}{@{}l}{\textit{Learning and memory}} \\
        Weight updates & $\times$ & $\times$ & $\times$ \\
        Guideline Memory & $\times$ & $\times$ & $\checkmark$ \\
        Strategy Memory & $\times$ & $\times$ & $\checkmark$ \\
        Joint memory update interval, $K$ (steps) & n/a & n/a & 2 \\
        \bottomrule
    \end{tabularx}
    \par\smallskip
    \begin{minipage}{\linewidth}
    \footnotesize
    $\checkmark$: enabled; $\times$: disabled; ---: not reported; n/a: not applicable.
    The original paper also reports 100-step inference runs. Code reference: \href{https://github.com/ypwang61/ThetaEvolve/tree/7c12898f5d7627af403e3aca643af76199fb5c20}{ThetaEvolve, commit \texttt{7c12898f}}.
    \end{minipage}
\end{table}

\clearpage
\subsection{TTT-Discover Framework}
\label{app:ttt-setup}

We evaluate Qwen3-8B and gpt-oss-120b on AC1, AC2, and Erd\H{o}s using
TTT-Discover~\citep{yuksekgonul2026learning}. Both models use the search
and evaluation settings below; Table~\ref{tab:ttt-setup} compares the
gpt-oss-120b settings with the original configuration.

\paragraph{Search and generation.}
PUCT selects eight parents from the reuse buffer per step, and the model
generates 64 candidates per parent. TTME allocates 32 to each generation
branch. We retain the released task prompts, construction-state reuse,
and verifier. Each rollout produces one complete Python program.
The buffer admits up to two children per parent and retains the initial states.

\paragraph{Completion and evaluation.}
Both models use vLLM with a 32,768-token context, a 50-token reserve,
temperature 1.0, top-$p$ 1.0, and no top-$k$ filtering.
For gpt-oss-120b, the released sampler allows 26,000 tokens for the prompt
and thinking, then continues in the final-answer channel when needed.
Qwen3-8B uses its chat template with an open \texttt{<think>} prefix and
the remaining context for the response, without forced final-answer continuation.
Programs have a 1,000-second runtime budget and a 1,100-second evaluator
timeout; failed programs receive no retry with error feedback.
Each run has 25,600 primary candidate attempts. Memory updates, strategy
selection, and adaptation require additional calls in w/ TTME.
\begin{table}[ht]
    \centering
    \small
    \caption{Settings for \texttt{gpt-oss-120b} on AC1, AC2, and Erd\H{o}s. Original denotes TTT-Discover with reinforcement learning~\citep{yuksekgonul2026learning}; both of our conditions use frozen weights. Counts are per run.}
    \label{tab:ttt-setup}
    \setlength{\tabcolsep}{4pt}
    \renewcommand{\arraystretch}{1.08}
    \begin{tabularx}{\linewidth}{@{}>{\raggedright\arraybackslash}X
        >{\centering\arraybackslash}p{0.20\linewidth}
        >{\centering\arraybackslash}p{0.165\linewidth}
        >{\centering\arraybackslash}p{0.165\linewidth}@{}}
        \toprule
        Setting & Original TTT-Discover & \multicolumn{2}{c}{Ours} \\
        \cmidrule(l){3-4}
        & w/ TTT & w/o TTT & w/ TTME \\
        \midrule
        \multicolumn{4}{@{}l}{\textit{Search budget and reuse buffer}} \\
        Search steps & 50 & 50 & 50 \\
        Parents $\times$ candidates per parent & $8\times64=512$ & $8\times64=512$ & $8\times64=512$ \\
        Primary candidates per run & 25,600 & 25,600 & 25,600 \\
        Buffer capacity / PUCT coefficient $c$ & 1,000 / 1.0 & 1,000 / 1.0 & 1,000 / 1.0 \\
        Maximum admitted children per parent & 2 & 2 & 2 \\
        \midrule
        \multicolumn{4}{@{}l}{\textit{Generation}} \\
        Context window (tokens) & 32,768 & 32,768 & 32,768 \\
        Prompt + thinking limit (tokens) & 26,000 & 26,000 & 26,000 \\
        Reasoning effort & high & high & high \\
        Forced final-answer continuation & $\checkmark$ & $\checkmark$ & $\checkmark$ \\
        Output format & Full program & Full program & Full program \\
        \midrule
        \multicolumn{4}{@{}l}{\textit{Evaluation and serving}} \\
        Runtime / eval. timeout (s) & 1,000 / 1,100 & 1,000 / 1,100 & 1,000 / 1,100 \\
        One retry with failure feedback & $\times$ & $\times$ & $\times$ \\
        Model serving & Tinker & vLLM & vLLM \\
        \midrule
        \multicolumn{4}{@{}l}{\textit{Learning and memory}} \\
        Weight updates & $\checkmark$ & $\times$ & $\times$ \\
        LoRA rank & 32 & n/a & n/a \\
        Guideline Memory & $\times$ & $\times$ & $\checkmark$ \\
        Strategy Memory & $\times$ & $\times$ & $\checkmark$ \\
        Joint memory update interval, $K$ (steps) & n/a & n/a & 2 \\
        \bottomrule
    \end{tabularx}
    \par\smallskip
    \begin{minipage}{\linewidth}
    \footnotesize
    $\checkmark$: enabled; $\times$: disabled; n/a: not applicable.
    Code reference: \href{https://github.com/test-time-training/discover/tree/6c40e82dab9d5de7416ac873ad5cd3106084aaed}{TTT-Discover, commit \texttt{6c40e82d}}.
    \end{minipage}
\end{table}

\clearpage

\section{Memory Implementation Details}
\label{app:memory-implementation}

Each run starts with both memories empty. We use one modification
instruction per parent ($N=1$). Section~\ref{app:prompts} provides the
prompt templates for distillation, consolidation, and adaptation.

\subsection{Memory Overview}
\label{app:memory-Overview}

Full TTME enables both components in Table~\ref{tab:memory-comparison}. Each component ablation removes
one memory component and retains the other.

\begin{table}[htbp]
    \centering
    \caption{Construction and use of the two memories. We update both
    from evaluated search outcomes every $K=2$ steps.}
    \label{tab:memory-comparison}
\begingroup
\small
\setlength{\tabcolsep}{6pt}
\renewcommand{\arraystretch}{1.16}
\noindent
\begin{tabularx}{\linewidth}{@{}>{\raggedright\arraybackslash}p{0.16\linewidth}
    >{\raggedright\arraybackslash}X
    >{\raggedright\arraybackslash}X@{}}
    \toprule
    \textbf{Aspect} & \textbf{Guideline Memory} & \textbf{Strategy Memory} \\
    \midrule
    Input evidence &
    Valid outcomes and execution or validity failures collected across parents. &
    Evaluated parent--child pairs in which the child is valid and improves on the parent. \\
    \addlinespace[7pt]
    Model input &
    Outcome summaries and the current guidelines. &
    Parent and child code, their diff, and evaluation results. \\
    \addlinespace[7pt]
    Extracted content &
    An operational lesson: a trigger, a mechanism, and a concrete action. &
    An experience note: the algorithmic idea expressed by a change and its application conditions. \\
    \addlinespace[7pt]
    Memory update &
    Merge new lessons with current guidelines, combine duplicates, and retain at most 24. &
    Group notes into strategy families. Create an entry for a new family, or consolidate new notes with the existing entry. \\
    \addlinespace[7pt]
    Use during generation &
    Include current guidelines in the prompt for both generation branches. &
    Select a strategy for the parent, then adapt its entry into a modification instruction for Strategy-Guided Generation. \\
    \bottomrule
\end{tabularx}\par
\endgroup

\end{table}

\subsection{Memory Updates}

We update both memories every $K=2$ search steps using outcomes
collected since the previous update.

\paragraph{Guideline Memory.}

\begin{enumerate}
    \setlength{\itemsep}{4pt}
    \setlength{\parskip}{0pt}
    \item \textbf{Summarize recent outcomes.}
    The outcome summary covers recent attempts across parents, including
    valid results, their scores, and execution or validity failures.

    \item \textbf{Distill guidelines.}
    The same fixed-weight base model reads the outcome summary and current
    guidelines, then proposes up to three lessons per update, at most
    one for failure avoidance. Each lesson specifies a trigger, explains
    the mechanism, and gives a concrete action. The parser extracts the
    lesson text from the response.

    \item \textbf{Match and merge.}
    For each proposed lesson, we compute its Jaccard similarity to stored
    guidelines using lowercased word-token sets. A similarity of at
    least $0.5$ counts as a match: we retain the stored text, increment
    its recurrence count, and refresh its last reaffirmed step. An
    unmatched lesson enters memory with count one and the current step.

    \item \textbf{Remove stale entries and enforce capacity.}
    Guidelines expire when more than 20 search steps have passed since
    their last reaffirmation. Among the remaining entries, we retain at
    most 24, ranked by recurrence count with recency breaking ties.

    \item \textbf{Include guidelines in generation prompts.}
    Both Direct Generation and Strategy-Guided Generation receive the
    retained guidelines as shared context. Each prompt displays up to
    240 characters per guideline; memory stores the full lesson text.
\end{enumerate}

\clearpage
\paragraph{Strategy Memory.}
\begin{enumerate}
    \setlength{\itemsep}{4pt}
    \setlength{\parskip}{0pt}
    \item \textbf{Select valid improvement cases.}
    We select evaluated parent--child pairs in which the child passes the
    task verifier and improves on its parent. Both scores must be finite;
    improvement follows the task's objective direction, consistent with
    $s(\text{child})>s(\text{parent})$ in Section~\ref{sec:strategy-memory}.

    \item \textbf{Extract experience notes.}
    The same fixed-weight base model reads the parent and child programs,
    their code diff, and evaluation results. It writes an experience
    note describing the algorithmic idea behind the change and
    its application conditions. Each note retains a reference to its
    source case; a change containing independent ideas can yield
    separate notes.

    \item \textbf{Group notes into strategy families.}
    The model compares new notes with existing strategy entries and
    groups notes that share an underlying algorithmic idea into the same
    family. It creates a new family for a distinct idea and preserves
    implementation differences as variants within each family.

    \item \textbf{Consolidate and store strategy entries.}
    For each family receiving new notes, the model combines those notes
    with the previous entry, if one exists, to update the description of
    the shared idea, application conditions, and implementation variants.
    We store the entry, its summary, and the supporting notes with source
    references. Families without new notes retain their existing entries.
\end{enumerate}

\subsection{Memory Hyperparameters}

The limits on guideline count and text length are configurable
hyperparameters. Table~\ref{tab:memory-hyperparameters} lists our settings
and candidate allocation.

\begin{table}[ht]
    \centering
    \small
    \caption{Memory hyperparameters and candidate allocation. We measure
    $K$ in search steps; $N$ is the number of modification
    instructions per parent. Character budgets apply to prompt-visible
    text.}
    \label{tab:memory-hyperparameters}
    \renewcommand{\arraystretch}{1.15}
    \begin{tabular*}{\linewidth}{@{\extracolsep{\fill}}lrr@{}}
        \toprule
        Setting & TTT-Discover & ThetaEvolve \\
        \midrule
        Joint memory update interval, $K$ (steps) & 2 & 2 \\
        Maximum stored guidelines & 24 & 24 \\
        Maximum proposed guidelines per update & 3 & 3 \\
        Guideline expiry window (search steps) & 20 & 20 \\
        Guideline matching threshold (Jaccard) & 0.5 & 0.5 \\
        Display budget per guideline (characters) & 240 & 240 \\
        Modification instruction budget (characters) & 700 & 700 \\
        \midrule
        Candidates per parent, $B$ & 64 & 16 \\
        Modification instructions per parent, $N$ & 1 & 1 \\
        Direct Generation candidates & 32 & 8 \\
        Strategy-Guided Generation candidates & 32 & 8 \\
        Candidates per modification instruction & 32 & 8 \\
        \bottomrule
    \end{tabular*}
\end{table}

\clearpage
\begingroup
\raggedbottom
\section{Prompts}
\label{app:prompts}
\label{app:prompt-templates}
\label{app:memory-output-examples}

The following prompts cover five memory operations. Section~\ref{app:prompt-inputs}
provides the input layouts supplied by the caller. Prompt wording
retains the original terms \emph{memo}, \emph{method memory}, and
\emph{direction}; subsection titles follow the Method terminology.
The example outputs were regenerated with \texttt{gpt-oss-120b} from historical
AC1 and AC2 inputs. Examples 1--3 are excerpts, and Example 5 is complete.
The guideline label is standardized, and \texttt{[...]} marks an omission.

\subsection{Guideline Memory Update}
\label{app:pair-guideline-update}

The caller appends the current guidelines and a summary containing the objective
direction, best score so far, representative successes across parents, and
available failure summaries. The template below shows the base instruction.

\begin{memoryprompt}{1}{}
You maintain a compact Guideline Memory of transferable lessons for an iterative program-search over a numerical construction problem. From the step outcomes you are given, distill AT MOST 3 guidelines \textemdash{} fewer, including none, when the outcomes support fewer. A guideline must CHANGE what the program does; if the failing code already satisfies your action, the action is not the fix and repeating it will reproduce the failure.\par
Each guideline must be ONE line in exactly this form:\par
{\ttfamily\fontsize{9}{10.5}\selectfont Guideline \textless{}n\textgreater{}: \textless{}trigger\textgreater{} -\textgreater{} \textless{}mechanism\textgreater{} -\textgreater{} \textless{}action\textgreater{}\par}
Number the guidelines consecutively from 1. Guidelines may be scientific what-drives-score lessons or failure-avoidance lessons. At most one failure-avoidance guideline. The action must be a concrete, executable instruction (an exact quantity, function, or step), never a vague suggestion. Do not quote these instructions back; emit only the guideline lines inside a \textless{}guidelines\textgreater{}...\textless{}/guidelines\textgreater{} block.
\end{memoryprompt}

\begin{memoryexample}{1}{}
Guideline 1: normal() got an unexpected keyword argument 'dtype' $\rightarrow$
\texttt{np.random.normal} does not accept a \texttt{dtype} parameter $\rightarrow$
replace each \texttt{np.random.normal(..., dtype=DTYPE)} with
\texttt{np.random.normal(...).astype(DTYPE)} (or simply omit the
\texttt{dtype} argument).
\end{memoryexample}

\clearpage
\subsection{Strategy Memory: Experience Note Extraction}
\label{app:pair-experience-extraction}

The caller appends the complete parent and child programs, their scores
formatted to ten decimal places, and their code diff.

\begin{memoryprompt}{2}{}
You are reading one step of an automated program-search run on a numerical optimisation problem. A parent program was modified into a child program, and the child scored better.
\par
Answer these four questions about THIS pair, briefly and concretely:
\begin{enumerate}[label=\arabic*.,leftmargin=1.6em,itemsep=3pt,topsep=3pt,parsep=0pt]
\item 
What substantive algorithmic change does the child make relative to the parent? Describe the change itself, not the names of functions.
\item 
Where in the program does the change take effect \textemdash{} which loop or phase?
\item 
Which part of the change is most plausibly connected to the score change?
\item 
Write ONE short memo (2-4 sentences) that would help someone making the next modification. Do not refer to any function or variable by name; a different implementation of the same idea must still match your memo.
\end{enumerate}
\par
Most parent-to-child modifications make ONE change. If this modification genuinely bundles two or three INDEPENDENT changes \mbox{-{}-} each would work, and could have been made, without the others \mbox{-{}-} answer all four questions for each change separately, and separate the answer blocks with a line containing only ===. Interlocking parts of one idea (a new stage plus the wiring that feeds it) are ONE change, not two.
\par
Write plain prose. Do not use section headers. Do not speculate beyond what the two programs show.
\end{memoryprompt}

\begin{memoryexample}{2}{}
\begin{enumerate}[label=\arabic*.,leftmargin=1.6em,itemsep=5pt,topsep=2pt,parsep=0pt]
\item When multiple convolution entries share the current maximum, the child
forms the descent direction by averaging the individual gradients for
all those active indices, whereas the parent simply picks the single
argmax entry.
\item This averaging is performed after the convolution is obtained and
before the step-size scaling, inside the inner descent loop.
\end{enumerate}
\texttt{[...]}
\begin{enumerate}[label=\arabic*.,start=4,leftmargin=1.6em,itemsep=5pt,topsep=2pt,parsep=0pt]
\item \textbf{Memo:} if the max convolution value occurs at several positions,
combine their gradients (e.g. by averaging) to obtain the sub-gradient
instead of using a single arbitrary index.
\end{enumerate}
\end{memoryexample}

\clearpage
\subsection{Strategy Memory: Strategy Entry Consolidation}
\label{app:pair-strategy-consolidation}

The caller places an existing-entry or new-entry prefix before the shared
instruction, then appends the corresponding inputs. The family label uses
its quoted string representation.

\begin{memoryprompt}{3}{}
Write ONE method memory for this family, with these four parts:
\par
\begin{enumerate}[label=\arabic*.,leftmargin=1.6em,itemsep=3pt,topsep=3pt,parsep=0pt]
\item 
MECHANISM - what the method actually does, in its own terms.
\item 
OBSERVED EXECUTION MODES - every distinct way it has been applied. Two modes are distinct when they differ in WHERE they run (before the loop, inside it, replacing the update, after it), in their SCHEDULE, in how the WORKING SET is chosen, or in their UPDATE / STATE-RESET semantics. They are the SAME mode when they differ only in wording or in ordinary numeric settings such as a step size or an iteration count.
\item 
WHEN EACH MODE SUITS A PROGRAM - the structural condition that makes a mode applicable to the program being improved.
\item 
WHAT MAY BE ADAPTED - which parts a per-parent rewrite is free to change.
\end{enumerate}
\par
Rules:
\begin{itemize}[label=-,leftmargin=1.6em,itemsep=3pt,topsep=3pt,parsep=0pt]
\item 
Do NOT restate the same execution mode more than once, however differently the evidence worded it.
\item 
Do NOT delete a genuinely distinct mode to be brief. There is no length limit; a simple family may need two sentences and a complex one a long paragraph.
\item 
Do NOT invent a numeric threshold or hard rule that the evidence does not support.
\item 
Output the memory only. No preamble, no JSON, no headings beyond the four parts above.
\end{itemize}
\end{memoryprompt}

\begin{memoryexample}{3}{}
OBSERVED EXECUTION MODES
\begin{enumerate}[label=\arabic*.,leftmargin=1.6em,itemsep=7pt,topsep=3pt,parsep=0pt]
\item \textbf{Post-up-sampling L-BFGS-B + final Adam polish} \textendash{}
The quasi-Newton refinement is called immediately after the two up-sampling
loops (which already produced a high-resolution candidate) and is followed
directly by a brief Adam polish; no further up-sampling occurs.
\item \textbf{Post-up-sampling + fine-tune L-BFGS-B $\rightarrow$ extra up-sampling
$\rightarrow$ final Adam polish} \textendash{}
The refinement is inserted after the up-sampling and the preceding Adam
fine-tuning stage, then its output is fed into an additional up-sampling step
before the concluding Adam polish.
\end{enumerate}
\end{memoryexample}

\clearpage
\subsection{Strategy-Guided Search: Strategy Selection}
\label{app:pair-strategy-selection}

In this template, \texttt{\{k\}} corresponds to the number of strategies per
parent, $N$, rather than the update interval $K$. The caller requests the
smaller of $N$ and the number of available strategy families. The program
and available-methods blocks are always included; other input blocks appear
only when their context is available. Score history includes the current
program and up to six recent ancestors, and each strategy summary contains
at most 400 characters.

\begin{memoryprompt}{4}{}
You are steering an automated program-search run on a numerical optimisation problem. Below is the program you must improve next, how it and its recent ancestors scored, what has already been tried on this line of descent, and a list of METHODS this run has already found to work somewhere.
\par
Select up to \{k\} methods from the list most likely to IMPROVE this program's score \textemdash{} including methods the program does not yet contain: similarity to what the program already does is not by itself a reason to choose or to avoid a method. The \{k\} must be DIFFERENT methods that lead to \{k\} different changes to this program: do not choose two methods that would make the same change. Take account of what this line of descent has already been given: repeating something that was tried here and did not move the score is usually worse than trying something this program has not seen. If several methods are equally plausible, prefer ones not already assigned to another group this step.
\par
Return ONLY a JSON object, no prose before or after, with the ids in order of preference:
\par
{\ttfamily\fontsize{9}{10.5}\selectfont \{"method\_ids": ["\textless{}id from the list\textgreater{}", "\textless{}id from the list\textgreater{}"]\}\par}
\end{memoryprompt}

\clearpage
\subsection{Strategy-Guided Search: Strategy Adaptation}
\label{app:pair-strategy-adaptation}

The caller appends the parent program and the selected strategy's full entry.
It omits the execution-summary block when unavailable and uses the strategy's
short direction only when the full entry is empty.

\begin{memoryprompt}{5}{}
You are steering an automated program-search run on a numerical optimisation problem. A method has been chosen for the program below. You are given the accumulated memory of that method: what it does, every distinct way it has been applied, when each way suits a program, and what may be adapted.
\par
Write a direction for THIS specific program: 2-4 sentences saying what to change and what would count as having done it. Pick the way of applying the method that suits the structure this program actually has \mbox{-{}-} where it runs, its schedule, how its working set is chosen \mbox{-{}-} and commit to that one choice rather than listing alternatives.
\par
The direction is the only thing the next author will see. Do not name any function or variable, do not mention scores or this run, and do not mention that a method list or a memory exists. Output the direction only \mbox{-{}-} plain prose, no preamble, no JSON, no headings.
\end{memoryprompt}

\begin{memoryexample}{5}{}
Insert a bounded L-BFGS-B optimisation step immediately after the second
up-sampling and its fine-tuning, using the exact gradient of the negative
lower-bound and enforcing non-negativity box constraints, and run it for a
modest iteration budget (e.g., a few hundred steps) until the convergence
tolerance is met. Then apply a brief Adam polishing pass (a few thousand steps)
to re-apply the projection and clean any residual noise. The program is complete
when the final candidate passed to the scorer is the result of this L-BFGS-B
followed by the short Adam refinement.
\end{memoryexample}

\clearpage
\subsection{Input Templates}
\label{app:prompt-inputs}

The caller fills the placeholders below and appends each input block to its
corresponding fixed instruction. Prompt 3 also receives a prefix before the
instruction; its two variants are shown below.

\begin{memoryinput}{1}
\#\# CURRENT playbook\par
\{current\_guideline\_memory\}\par
\smallskip
\#\# THIS STEP\par
\{step\_outcome\_summary\}\par
\smallskip
Emit the updated lessons now.\par
\end{memoryinput}

\begin{memoryinput}{2}
\#\# PARENT PROGRAM (score \{parent\_score\})\par
\textasciigrave{}\textasciigrave{}\textasciigrave{}python\par
\{parent\_program\}\par
\textasciigrave{}\textasciigrave{}\textasciigrave{}\par
\smallskip
\#\# CHILD PROGRAM (score \{child\_score\}, better)\par
\textasciigrave{}\textasciigrave{}\textasciigrave{}python\par
\{child\_program\}\par
\textasciigrave{}\textasciigrave{}\textasciigrave{}\par
\smallskip
\#\# DIFF (parent -\textgreater{} child)\par
\textasciigrave{}\textasciigrave{}\textasciigrave{}diff\par
\{parent\_to\_child\_diff\}\par
\textasciigrave{}\textasciigrave{}\textasciigrave{}\par
\end{memoryinput}

\begin{memoryinput}{3}
{\normalfont\small\bfseries Existing entry: prefix before Prompt 3\par}\smallskip
Here is the current method memory for the family \{family\_label\_repr\}, and the notes that arrived at this boundary. Rewrite the memory so it covers both. Keep every execution mode already in it unless a new note shows it was wrong.\par
\medskip
{\normalfont\small\bfseries Inputs after Prompt 3\par}\smallskip
\#\# CURRENT MEMORY\par
\{current\_strategy\_entry\}\par
\smallskip
\#\#\# new note 1\par
\{experience\_note\_1\}\par
\smallskip
\#\#\# new note 2\par
\{experience\_note\_2\}\par
\smallskip
...\par
\medskip
{\normalfont\small\bfseries New entry: alternative prefix before Prompt 3\par}\smallskip
You are starting the method memory for the family \{family\_label\_repr\} from nothing.\par
\medskip
{\normalfont\small For a new entry, append only the numbered new notes and omit the \texttt{CURRENT MEMORY} block. The number of notes varies.\par}
\end{memoryinput}

\begin{memoryinput}{4}
\#\# PROGRAM TO IMPROVE\par
\{parent\_program\}\par
\smallskip
\#\# WHAT THIS PROGRAM ACTUALLY EXECUTED (self-reported)\par
\{self\_reported\_execution\_summary\}\par
\smallskip
\#\# SCORE OF THIS PROGRAM AND ITS RECENT ANCESTORS\par
\{Higher\_or\_LOWER\} is better. \{scores\_most\_recent\_first\} (most recent first)\par
\smallskip
\#\# WHAT THIS LINE OF DESCENT HAS ALREADY BEEN GIVEN\par
\{lineage\_assignment\_history\}\par
\smallskip
\#\# WHAT HAS GONE WRONG BEFORE\par
\{memory\_context\}\par
\smallskip
\#\# METHODS ALREADY ASSIGNED TO OTHER GROUPS THIS STEP\par
- \{already\_assigned\_method\_1\}\par
- \{already\_assigned\_method\_2\}\par
...\par
\smallskip
\#\# METHODS THIS RUN HAS FOUND\par
- id \textasciigrave{}\{method\_id\_1\}\textasciigrave{}: \{strategy\_summary\_1\}\par
- id \textasciigrave{}\{method\_id\_2\}\textasciigrave{}: \{strategy\_summary\_2\}\par
...\par
\end{memoryinput}

\begin{memoryinput}{5}
\#\# PROGRAM TO IMPROVE\par
\{parent\_program\}\par
\smallskip
\#\# WHAT THIS PROGRAM ACTUALLY EXECUTED (self-reported)\par
\{self\_reported\_execution\_summary\}\par
\smallskip
\#\# THE CHOSEN METHOD'S ACCUMULATED MEMORY\par
\{strategy\_label\_or\_id\}\par
\smallskip
\{full\_strategy\_entry\_or\_fallback\_direction\}\par
\end{memoryinput}

\clearpage
\endgroup

\section{Discovered Constructions}
\label{app:discovered-constructions}

Figure~\ref{fig:banked-constructions} shows constructions for the five tasks defined in Appendix~\ref{app:task-definitions}.

% Requires graphicx, amsmath and subcaption.
% Five model-first prose captions; two continued figures preserve labels (a)-(e).
\begin{figure}[!htbp]
\centering
\captionsetup[subfigure]{font=small,labelfont=bf,skip=9pt,justification=raggedright,singlelinecheck=false}
\begin{subfigure}[t]{\linewidth}
  \centering
  \includegraphics[width=\linewidth]{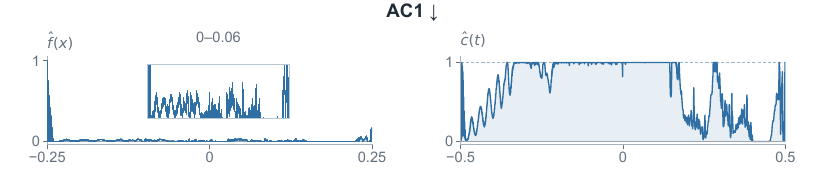}
  \caption{\textit{gpt-oss-120b} produced a 32\,000-bin AC1 construction with an objective value of $1.502745082$. The plots show the normalized step function $\hat f(x)$ (left) and autoconvolution $\hat c(t)$ (right).}
  \label{fig:banked-ac1}
\end{subfigure}
\par\vspace{5mm}
\begin{subfigure}[t]{\linewidth}
  \centering
  \includegraphics[width=\linewidth]{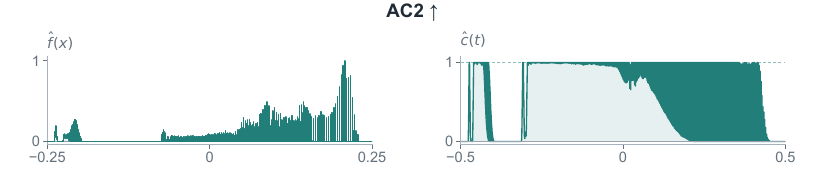}
  \caption{\textit{gpt-oss-120b} produced a 65\,536-bin AC2 construction with an objective value of $0.960784$. The plots show the normalized step function $\hat f(x)$ (left) and autoconvolution $\hat c(t)$ (right).}
  \label{fig:banked-ac2}
\end{subfigure}
\par\vspace{5mm}
\begin{subfigure}[t]{\linewidth}
  \centering
  \includegraphics[width=\linewidth]{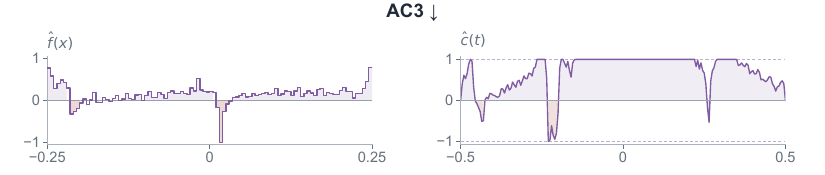}
  \caption{\textit{Distill-Qwen3-8B} produced a 100-bin AC3 construction with an objective value of $1.464224300$. The plots show the normalized step function $\hat f(x)$ (left) and autoconvolution $\hat c(t)$ (right).}
  \label{fig:banked-ac3}
\end{subfigure}
\par\vspace{5mm}
\begin{subfigure}[t]{\linewidth}
  \centering
  \includegraphics[width=\linewidth]{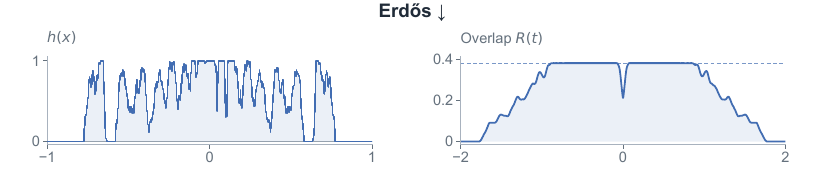}
  \caption{\textit{gpt-oss-120b} produced the 801-bin Erd\H{o}s construction with a maximum overlap of $0.380861541$. The left plot shows its density $h(x)$ on $[-1,1]$, and the right plot shows the corresponding overlap $R(t)$.}
  \label{fig:banked-erdos}
\end{subfigure}
\caption{Discovered constructions for AC1--AC3 and Erd\H{o}s. The AC1 inset shows normalized heights 0--0.06. Arrows indicate the direction of improvement.}
\label{fig:banked-constructions}
\end{figure}
\clearpage
\begin{figure}[!htbp]
\ContinuedFloat
\centering
\captionsetup[subfigure]{font=small,labelfont=bf,skip=9pt,justification=raggedright,singlelinecheck=false}
\begin{subfigure}[t]{\linewidth}
  \centering
  \includegraphics[width=\linewidth]{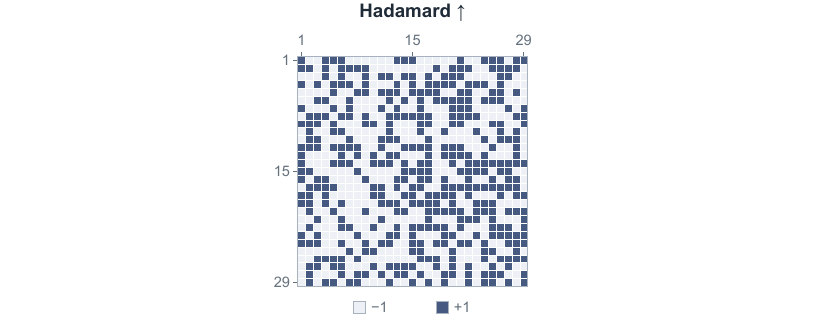}
  \caption{\textit{Distill-Qwen3-8B}, with memory enabled, produced the $29\times29$ sign matrix $H$, attaining an objective value of $0.579938205$. Dark and light cells denote $+1$ and $-1$, respectively.}
  \label{fig:banked-hadamard}
\end{subfigure}
\caption{Discovered constructions (continued). Panel (e) preserves the original row and column order of the Hadamard construction.}
\label{fig:banked-constructions-continued}
\end{figure}

\end{document}